\documentclass{article}
\usepackage[utf8]{inputenc}
\usepackage{authblk}
\usepackage{setspace}
\usepackage[margin=1.25in]{geometry}
\usepackage{graphicx}
\graphicspath{ {./figures/} }
\usepackage{subcaption}
\usepackage{amsmath}
\usepackage{hyperref}
\hypersetup{
    colorlinks=true,
    citecolor = black
}

\RequirePackage{listings}
\RequirePackage[en-GB]{datetime2}
\DTMlangsetup{showdayofmonth=false}
\RequirePackage{xcolor}
\RequirePackage{microtype}  %
\RequirePackage{xurl}  %

\lstdefinestyle{mystyle}{
    backgroundcolor=\color{backcolour},
    commentstyle=\color{codegreen},
    keywordstyle=\color{magenta},
    numberstyle=\tiny\color{codegray},
    stringstyle=\color{codepurple},
    basicstyle=\ttfamily\footnotesize,
    breakatwhitespace=false,
    breaklines=true,
    captionpos=b,
    keepspaces=true,
    numbers=left,
    numbersep=5pt,
    showspaces=false,
    showstringspaces=false,
    showtabs=false,
    tabsize=2
}
\usepackage[T1]{fontenc}
\usepackage[british]{babel}
\usepackage{csquotes}
\usepackage{pdflscape}

\usepackage{xspace}
\usepackage{algorithm}
\usepackage{algorithmicx}
\usepackage{algpseudocode}
\usepackage{float}
\usepackage{multirow}
\usepackage{amssymb}
\usepackage{adjustbox}

\newcommand{\product}{EcoMapper\xspace}
\newcommand{\metric}{CI\xspace}

\def\fillandplacepagenumber{%
    \par\pagestyle{empty}%
    \vbox to 0pt{\vss}\vfill
    \vbox to 0pt{\baselineskip0pt
    \hbox to\linewidth{\hss}%
    \baselineskip\footskip
    \hbox to\linewidth{%
        \hfil\thepage\hfil}\vss}}

\usepackage[backend=biber,style=numeric,sorting=none]{biblatex}
\title{Segmentation of arbitrary features in very high resolution remote sensing imagery}

\author[1,2,*]{Henry Cording}
\author[1]{Yves Plancherel}
\author[1]{Pablo Brito-Parada}

\affil[1]{Department of Earth Science and Engineering, Imperial College London, London SW7 2AZ, United Kingdom}
\affil[2]{Vattenfall Europe Information Services GmbH, Berlin 10829, Germany}
\affil[*]{Address correspondence to: henry.cording@pm.me}

\date{}

\begin{document}

\maketitle

\begin{abstract}
Very high resolution (VHR) mapping through remote
sensing (RS) imagery presents a new opportunity to inform decision-making and sustainable
practices in countless domains.
Efficient processing of big VHR data requires automated tools applicable to numerous
geographic regions and features.
Contemporary RS studies address this challenge by employing deep learning (DL) models for specific datasets or features, which limits their applicability across contexts.

The present research aims to overcome this limitation by introducing \product, a scalable solution to segment
arbitrary features in VHR RS imagery.
\product fully automates processing of geospatial data, DL model training, and inference.
Models trained with \product successfully segmented two distinct features in a
real-world UAV dataset, achieving scores competitive with prior studies which employed context-specific
models.

To evaluate \product, many additional models were trained on permutations of principal field survey
characteristics (FSCs). %
A relationship was discovered allowing derivation of optimal ground sampling distance from
feature size, termed Cording Index (\metric).
A comprehensive methodology for field surveys was developed to ensure DL methods can be applied
effectively to collected data.

The \product code accompanying this work is available at \href{https://github.com/hcording/ecomapper}{this https URL}.
\end{abstract}

\section{Introduction}\label{sec:introduction}
Over the last four decades, the spatial resolution of both satellite and unmanned aerial
vehicle (UAV) imagery has improved rapidly~\cite{Coffer2020}, reaching 30--50 cm/pixel for
privately owned satellites~\cite{FotsoKamga2021,dakir2021optical}, and less than 1
cm/pixel for UAVs~\cite{McKellar_Shephard_Chabot_2020,AncinMurguzur2019,Casella2016}.
These developments enable species-specific
recognition~\cite{Ferreira2020,Kattenborn2020} and feature identification at up to centimetre
scale~\cite{Neupane2019,Tetila2020,jay2019}.
Concurrently, increases in spatial and temporal resolution call for new data processing
capabilities~\cite{Zhang2016}.
As a result, machine learning and in particular deep learning (DL) methods have seen several
adaptations for automated analysis of remote sensing (RS) imagery in recent
years~\cite{ball2017,lei2019,Youssef2020}.

\begin{figure}[ht]
    \centering
    \includegraphics[width=.97\textwidth]{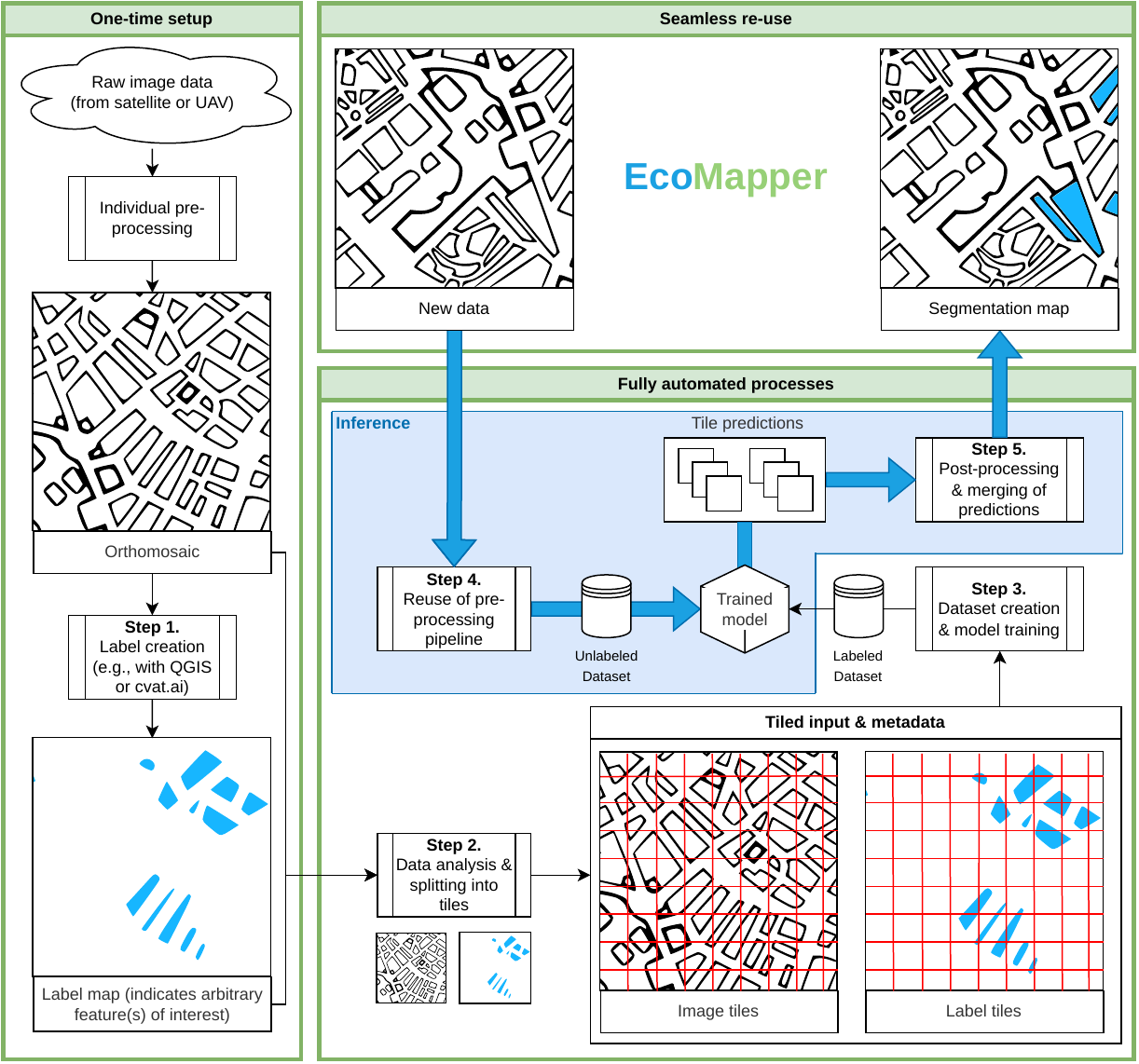}
    \caption{
        A high level overview of \product's architecture.
        Data pre- and post-processing, as well as model training, evaluation, and inference
        are fully automated.
    }
    \label{fig:architecture}
\end{figure}
\newpage
With the availability of very high resolution (VHR) RS imagery and increasingly powerful DL models,
a novel opportunity emerges to map a
multitude of features, such as insects and fauna, at centimeter scale across entire landscapes.
This would facilitate biodiversity monitoring at very high temporal resolution,
to oversee endangered species and support their rehabilitation.

In the context of climate change, VHR mapping could provide detailed insights into changes in land
cover, vegetation health, and water resources, which are critical for understanding
the impacts of climate change~\cite{Pielke2016, Sleeter2018, Yigal2014, Mall2006}.
VHR mapping could also enable resource valuation and management on an unprecedented scale,
to aid in establishing sustainable practices such as precision agriculture.

In sum, the development of robust
systems for VHR mapping could prove invaluable for informing decision-making and
sustainable development globally.

\subsection{Problem Description}\label{subsec:problem-description}
A major issue in contemporary RS research preventing large scale VHR mapping is the strong
fluctuation in how DL models are designed and used.
Many studies focus on building custom models to
perform well on particular datasets or features, visible clearly in surveys on the use
of DL in RS~\cite{ball2017, lei2019, Youssef2020}.
Literature review additionally revealed a conspicuous lack of standardized DL
workflows~\cite{Ferreira2020, Kattenborn2020, Neupane2019, Tsouros2019, Barbedo2019}.

This makes results hard to reproduce, especially since code is rarely open
sourced~\cite{Heil2021, Machicao2022}, which
adds to the already high barrier of entry to leverage DL that arises from
model complexity and computational cost~\cite{ball2017}.
Most importantly, with researchers tailoring their own models to specific datasets,
transferability of implementations is hindered.
This is despite VHR RS imagery being extremely diverse, containing countless features
whose size and spectral properties can vary significantly.
It therefore cannot be feasible to design entire models and workflows anew for each
feature to be studied.

\subsection{A case for \product}\label{subsec:case-for-ecomapper}
To address the shortcomings of context-specific solutions, this research introduces \product\footnote{\url{https://github.com/hcording/ecomapper}}, a
DL-based application for semantic segmentation of arbitrary features in VHR RS imagery.
Designed to provide a reusable and configurable approach to feature segmentation, the aim of
\product is to reduce duplicate RS research efforts and aid the reproducibility of results, while
also democratizing access to DL technology.
An overview is provided in Fig.~\ref{fig:architecture}.

Semantic segmentation divides an image into segments, where each segment identifies an
object class (e.g., car, pedestrian, building).
Although semantic segmentation is not the only computer vision task in RS, along with Land
Use/Land Cover (LULC) classification it constitutes a significant portion of contemporary RS
research~\cite{lei2019}.
Notably, LULC classification can be addressed with semantic segmentation models~\cite{Hassan2020},
thus \product presents a generalizable solution to a large problem space in RS\@.

Solutions related to \product exist but are either not applicable
to RS data, or fail to eliminate the need for specific models to be designed on a per use case
basis.
They are discussed below.

\subsubsection{Raster Vision}\label{subsubsec:raster-vision}
Raster Vision~\cite{Azavea2023} is a library and framework for applying DL to RS imagery.
It supports several geospatial data formats and implements processing
routines to make RS data compatible with DL model training.
Raster Vision can be used as a library for training existing PyTorch models with
RS data, but also provides a framework to train prebuilt models on RS imagery.
There are several reasons why Raster Vision is not suitable for repeated application across
data- and feature-sets, which is the main goal of \product.

\textit{Model selection.}
Raster Vision provides only a limited selection of prebuilt models via the PyTorch
Hub~\cite{PyTorchHub}, which does not feature recent state-of-the-art models.
Recently published models thus need to be implemented or sourced manually.
As motivated in section~\ref{sec:introduction}, there is need for a solution in which the
choice of model is abstracted away.
Section~\ref{subsec:pipeline-development} will show that model training in \product
was integrated with the MMSegmentation library~\cite{mmseg2020}, making it trivial to update or swap
models, unlike in PyTorch (or similar libraries).

\textit{Ease of use.}
Raster Vision requires programming experience even for the simplest use cases.
Python script files need to be written to describe a particular task to the framework.
In contrast, \product was developed under a ready-to-use paradigm -- while it can easily be extended
programmatically, e.g., to use a different model for segmentation, the published version can
be run from the commandline without writing any code; and it is still
applicable to any dataset and feature(s) of interest.

\textit{Data labeling support.}
Raster Vision does not integrate with data labeling tools, which are an integral part
of segmentation workflows.
Section~\ref{subsec:labeling-methods} will detail measures taken to integrate \product
with various labeling tools and formats, providing a solution that covers the entire segmentation
workflow, and is accessible to users without prior labeling experience.

\subsubsection{Segment Anything}\label{subsubsec:segment-anything}
The Segment Anything Model (SAM)~\cite{kirillov2023segany} is a recent zero-shot
segmentation model developed by Meta (Facebook).
As the name suggests, the model is intended for segmenting arbitrary scenes without requiring
additional tuning.

SAM was trained on the very diverse Segment Anything dataset~\cite{kirillov2023segany}, with no
particular feature focus, to cover as many use cases as possible.
This makes it a robust solution for segmenting commonly occurring features.
The model can either be used interactively by clicking on image regions to include or exclude, or it
can segment a scene autonomously.

\product produces models which specialize on particular features of interest.
These models outperform SAM when studying features that are underrepresented in
day-to-day imagery.
As shown in appendix~\ref{sec:appendix:sam}, using SAM the segmentation of a unique feature
requires several guiding inputs from the user, and fails entirely in the autonomous mode.
The model trained by \product manages to segment the feature precisely and without aid.

Lastly, SAM does not provide any means for pre- or post-processing of geospatial
data, which is a key component of \product, and introduces non-trivial
challenges for training DL models.

\subsection{Challenges in applying \product across varying real-world contexts}\label{subsec:synergy}
A solution like \product is urgently required to cope with processing demands of large scale VHR data.
However, \product is not a holistic solution to arbitrary feature segmentation, due to
interdependency with field survey characteristics (FSCs).
FSCs dictate the amount and quality of data available for DL model training.
Feature size, image resolution, and survey extent affect DL model performance and the
cost, duration, and feasibility of field surveys.
A comprehensive surveying methodology must be developed to respect surveying resources and
ensure effective application of DL to collected data.

Impacts of FSCs on DL performance must be understood to build a DL-informed surveying method.
Prior studies selectively reviewed resolution, feature size, or dataset size as influencing factors
on model performance, but a combination of factors was not considered in depth.

In~\cite{Neupane2019} and~\cite{Velumani2021}, the impact of ground sampling
distance (GSD) on DL performance was described.
GSD measures the distance on Earth's surface covered by the width (or height) of a single pixel.
GSD has a large impact on model performance~\cite{Neupane2019, Velumani2021, Barbedo2019}, but
the impact is dependent on feature size~\cite{Brown2022} as larger features are more visible in low
resolution imagery.
In~\cite{Brown2022} the effect of feature size on performance was demonstrated, but without
discussing a general relationship between feature size and GSD\@.
Effects of survey extent and dataset size on DL performance in RS studies are
discussed even more sparingly~\cite{Guetter2022, Calhoun2022}.

\subsection{Research gaps \& objectives of this study}\label{subsec:research-objectives}
To summarize, the following research gaps emerge:

\begin{enumerate}
    \item DL models are applied to specific use cases.
    Introduction of VHR imagery and abundance of visible features call for a more widely applicable
    solution to feature extraction.
    \item A generalizable DL solution requires a surveying methodology that links FSCs to DL,
    ensuring applicability of DL to collected data.
\end{enumerate}

\noindent To address the prior shortcomings, the following objectives were posed for this study:
\begin{enumerate}
    \item To design \product for automatic segmentation of arbitrary features in VHR RS imagery,
    without requiring dataset-specific model tuning.
    \item To simulate principal FSCs, such as resolution, feature size, and
    survey extent, and evaluate their effects on model performance.
    \item To design a generalizable field survey workflow to accommodate for FSCs and the
    application of DL.
\end{enumerate}

\section{Methodology}\label{sec:methodology}
A real-world UAV dataset was first obtained (section~\ref{subsec:dataset-overview}) and two
distinct features were labeled (section~\ref{subsec:labeling-methods}).
\product was then developed (section~\ref{subsec:pipeline-development}) without specific
considerations to the dataset at hand.
Models were repeatedly trained with \product to simulate several permutations of FSCs
(section~\ref{subsec:simulation-of-survey-characteristics}).
Multiple degradation methods were considered to synthetically reduce GSD\@.
Lastly, experiments were set up and run on the HPC platform of Imperial College London
(section~\ref{subsec:model-training}).

\subsection{Dataset overview}\label{subsec:dataset-overview}
For this study, an RGB orthomosaic (perspective-normalized stitching of many, partially overlapping
images) was used.
The image mapped approximately 3 km$^2$ of the Sto.\ Niño legacy mining site in Tublay, Philippines.
An excerpt is depicted in appendix~\ref{sec:appendix:dataset}.
A DJI Mavic 3 MS drone equipped with a 20MP sensor was flown at 50 metre
altitude to capture this data, yielding a GSD of 0.022 m/px.
Images were taken in March 2023.

\subsection{Labeling methods and integration with \product}\label{subsec:labeling-methods}
\begin{figure}[ht]
    \centering
    \begin{subfigure}[b]{0.49\textwidth}
        \includegraphics[width=\textwidth]{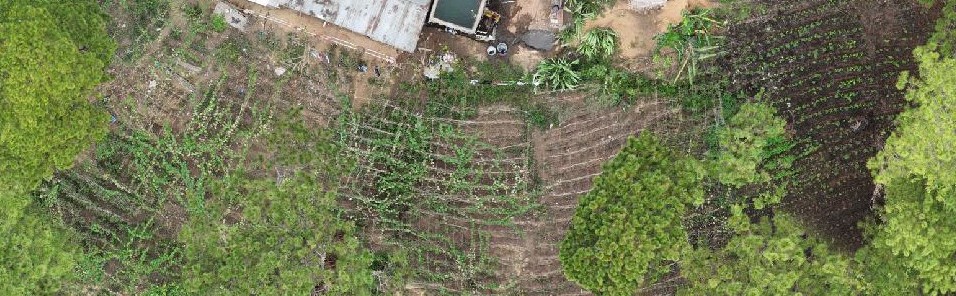}
        \caption{}
        \label{fig:chayote-example}
    \end{subfigure}
    \begin{subfigure}[b]{0.49\textwidth}
        \includegraphics[width=\textwidth]{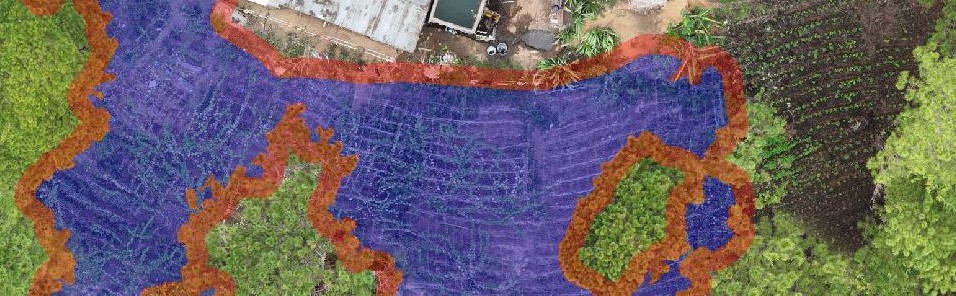}
        \caption{}
        \label{fig:labeled-chayote}
    \end{subfigure}

    \begin{subfigure}[b]{0.325\textwidth}
        \includegraphics[width=\textwidth]{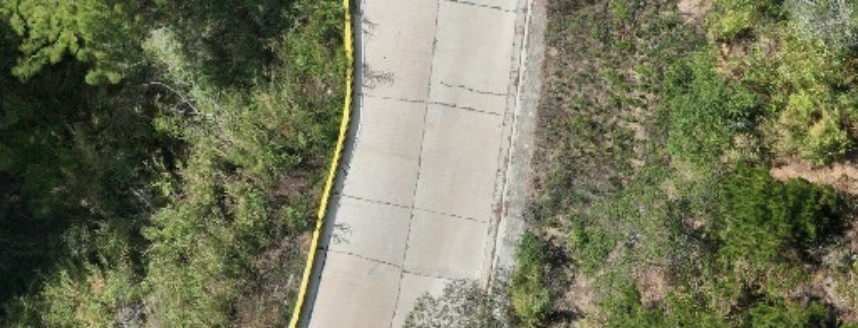}
        \caption{}
        \label{fig:road}
    \end{subfigure}
    \begin{subfigure}[b]{0.325\textwidth}
        \includegraphics[width=\textwidth]{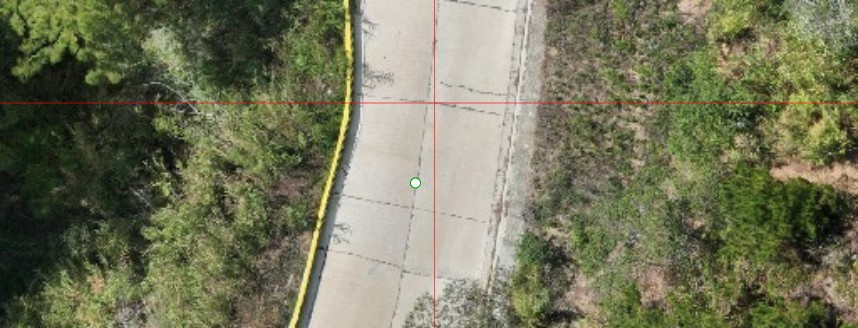}
        \caption{}
        \label{fig:road-with-sam-points}
    \end{subfigure}
    \begin{subfigure}[b]{0.325\textwidth}
        \includegraphics[width=\textwidth]{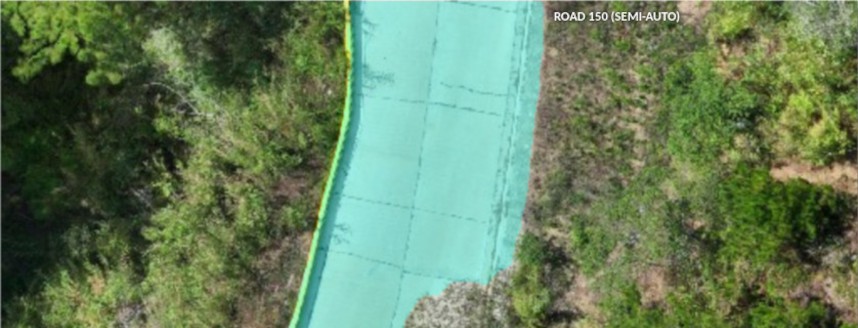}
        \caption{}
        \label{fig:road-sam-autolabel}
    \end{subfigure}
    \caption{Overview of the labeling process.
    Top row: QGIS labeling. (a) Partial view of a Chayote plantation in the Sto.\ Niño region; (b)
        an overlay of the manually drawn label map for Chayote,
        labels were palettized for visualization. Blue indicates ``Chayote'', red indicates ``Border''
        (uncertainty).
        Bottom row: CVAT labeling. (c) Input image; (d) points are placed indicating the feature to
        label; (e) the label (cyan) is generated automatically by CVAT.}
        \label{fig:labeling-process}
\end{figure}

Data were labeled manually using QGIS~\cite{qgis2023}, and
CVAT~\cite{CVAT2022} which employs DL models to accelerate labeling.
Fig.~\ref{fig:labeling-process} illustrates the labeling workflow.
In support of reproducible research, instructional videos were recorded explaining the labeling
procedure in QGIS\@.
\product is compatible with CVAT labels, and a video was recorded
explaining the use of CVAT and how to import results into \product.
Details and video URLs are documented in appendix~\ref{sec:appendix:labeling-tutorials}.

Chayote plantations (appendix~\ref{sec:appendix:chayote-plantation-example}) were chosen as initial
feature to label and segment.
A ``border'' class was introduced to indicate uncertainty w.r.t.\ where plantations end and the
background starts.
Roads were labeled (Fig.~\ref{fig:road}--\ref{fig:road-sam-autolabel})
to investigate whether \product could segment a feature with vastly different size and spectral
properties.
Roads did not have a border class as they were easily distinguished from the background.

\subsection{Development of the \product pipeline}\label{subsec:pipeline-development}
\product was developed as OS-independent application with a commandline interface (CLI) using
Python.
The project's code repository features thorough installation and usage instructions.
Sphinx~\cite{Sphinx2023} was used to organize code documentation into HTML format.
The \verb|pip-tools| library~\cite{Piptools2023} was utilized to generate an exhaustive 
file of pinned requirements, aiding build reproducibility.
Integration of the GDAL library~\cite{GDAL2023} to work with geospatial data was achieved with
Anaconda~\cite{Anaconda2023}, to avoid complications with OS-specific packages otherwise required by
GDAL\@.
All software used during EcoMapper`s development is free and open source.

\subsubsection{Architecture}\label{subsubsec:architecture}
Three components make up the pipeline illustrated in
Fig.~\ref{fig:architecture}: Data preprocessing, model training/evaluation, and
postprocessing of model predictions into a segmentation map.
\product is built up from ``Tasks'' to abstract these steps.
For example, model training is implemented through the \verb|TrainTask|.
In appendix~\ref{sec:appendix:product-task-network} the network of Tasks and their interdependencies
are depicted.
The CLI can either run the full pipeline (all tasks in sequence) or
run any Tasks in isolation, e.g., to prepare a dataset for future training on a different machine.

Several notable features were built into \product, including multiprocessing, support for a wide
range of hardware, and the ability to work with a multitude of geospatial image and label formats.
A complete description can be found in appendix~\ref{sec:appendix:notable-features-of-pipeline}.

\subsubsection{Preprocessing -- Data fragmentation and dataset creation}
Inputs to DL models are small, typically 512x512 pixels or less, but geospatial data are far larger.
Splitting RS imagery is challenging because orthomosaics typically do not fit into memory.
Prediction artifacts at tile boundaries can also arise since adjacent tiles are input separately to
the model.

Input images to \product are memory mapped (read incrementally from disk, instead of at once).
A sliding window is used to introduce overlap between tiles.
Blending tiles significantly increases prediction coherence, and provides $(1/s)^2$
times more data for model training, where $0<s<1$ is the stride ($s=0.5$ was chosen here).
The resulting image and label tiles are stored in compressed JPG format at 90\% quality and PNG
format (lossless compression), respectively.
Georeference information in the input is saved and later restored, so the predicted
segmentation map can again be imported into GIS software and
will be positioned correctly.

\subsubsection{MMSegmentation integration}\label{subsubsec:pipeline-model-training}
DL research advances rapidly.
For \product to be a future-proof solution, manual implementation of
semantic segmentation models from scratch was deemed unsuitable.

Instead, MMSegmentation~\cite{mmseg2020} was integrated to facilitate model training.
MMSegmentation is an actively maintained library for semantic segmentation, developed by
OpenMMLab~\cite{Openmmlab}.
The library has been used in many prior studies (
e.g.,~\cite{park2023multiearth, zhang2021knet, liang2022gmmseg}) and contains numerous
state-of-the-art DL models for semantic segmentation.
New models are added regularly as research progresses.
MMSegmentation centers around simple configuration files with a high degree of
abstraction, which result in the creation of PyTorch models in the backend.
Updating or replacing models is therefore simplified compared to direct use of PyTorch or other DL libraries.

\textit{Training strategies.}
\product extends the MMSegmentation training procedure to mitigate
class imbalance between features and the background.
The pipeline calculates the class distribution over the training dataset and assigns a weight to
each tile, as shown in appendix~\ref{subsec:appendix:sample-weights}.
These weights are used during training to oversample minority class(es) for each batch of
training data.

Samples are then subjected to randomized image
augmentations, using the Albumentations~\cite{Buslaev2020} library.
Augmentations such as downscaling and color shift help
reduce overfitting on the training set~\cite{Krizhevsky2012, Mikolajczyk2018, Shorten2019}.
By increasing the variety of tiles when oversampling minority classes,
the likelihood of having identical tiles in a batch is reduced.

\textit{Splitting data for training and evaluation.}
RS data are spatially auto-correlated~\cite{Karasiak2021}, meaning
the distance between pixels influences their similarity.
Thus, the assumption of independent and identically distributed (IID) random variables does not
hold, and regular DL approaches to shuffle and split data must be avoided.

\product offers two splitting methods.
The first method divides the dataset into train, validation, and test sets along the horizontal line.
After the split, only the train set is shuffled.
Method 1 addresses the problem of spatial leakage, but can lead to suboptimal proportions of
non-background samples in each set.
In the worst case, the training set only contains background samples, giving the model nothing to
train on.

Method 2 allows users to manually split data with QGIS\@.
The resulting sets with balanced class distribution can be passed to \product for automatic
model training and evaluation using three datasets.

\subsubsection{Postprocessing -- Merging model predictions}
As image tiles partially overlap, so do the model's predictions.
Merging of predictions is non-trivial and can heavily impact segmentation map quality.
Appendices~\ref{subsec:appendix:logit-merging} and ~\ref{subsec:appendix:crop-merging} outline
considered strategies, including visual comparison of merge results.

The first approach (logit-merge) merges overlapping tiles using the model's confidences (logits).
Method two (crop-merge) crops overlapping tiles, such that only the center of tiles is used (except
at image boundaries).
Crop-merging generates higher quality results and executes faster.

\subsection{Simulating survey characteristics}\label{subsec:simulation-of-survey-characteristics}
\begin{figure}[ht]
    \centering
    \includegraphics[width=\textwidth]{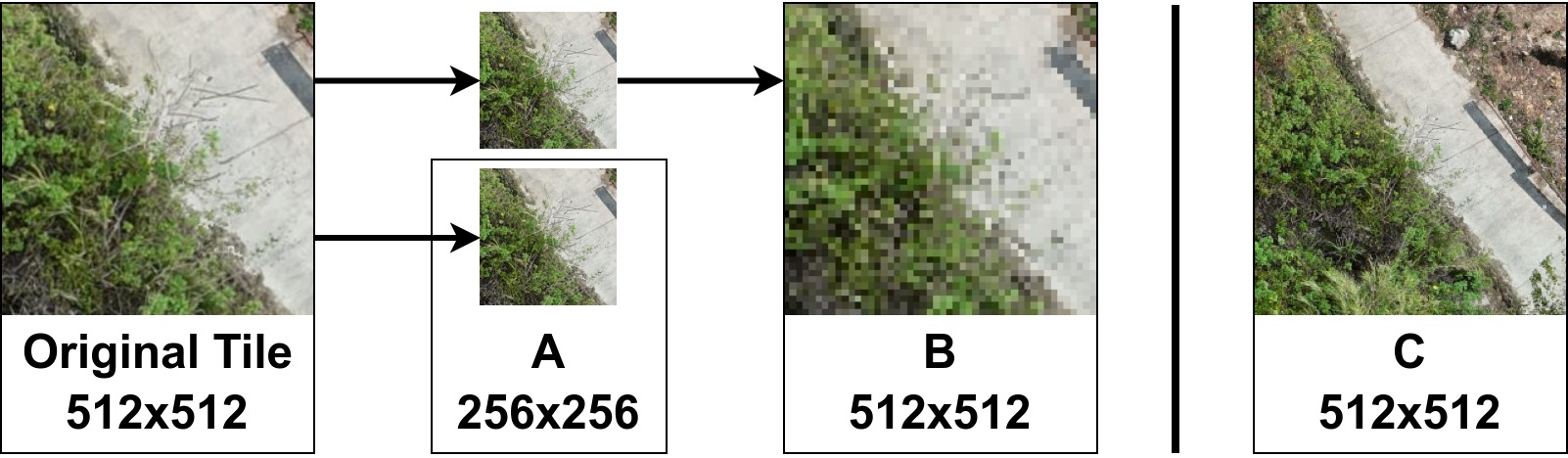}
    \caption{
        Methods of resolution degradation. The original image (top left) can be downsized (A),
        downsized and upscaled to the original tile dimensions (B), or the orignal orthomosaic can
        be downsized and split into tiles anew, yielding fewer tiles that cover more spatial
        distance and appear ``zoomed out'' (C).
    }
    \label{fig:resolution-degradation}
\end{figure}

As argued in section~\ref{sec:introduction}, a meaningful evaluation of \product had to consider
principal survey characteristics: feature size, GSD, and survey extent.
To study the effect of feature size, models were trained on Chayote plantations and roads
which have vastly different size (and spectral properties).

To study the impacts of GSD, per feature models were trained on 12 versions of the
dataset with decreasing GSD\@.
The 9 lowest GSDs were chosen from satellites capturing VHR imagery,
to gauge the viability of satellite data for VHR studies.
Appendix~\ref{sec:appendix:resolutions} lists all considered GSDs; due to computational constraints,
the highest considered resolution was 0.08 m/px.

Different synthetic degradation methods were applied to reduce GSD, as illustrated in
Fig.~\ref{fig:resolution-degradation}.
Appendix~\ref{sec:appendix:degradation-methods} describes each method in detail.
Note that method B only degrades images, while leaving labels at native quality.
Investigating multiple degradation methods is necessary to identify a general trend in performance
degradation, making the evaluation more robust to particularities of individual methods.

Lastly, models were trained on the original data with a decreasing number of images
in the training set, to simulate the effect of survey extent on model performance.
A total of 99 models were trained for this study.

\subsection{Model training}\label{subsec:model-training}

The Mask2Former~\cite{cheng2021mask2former} model was chosen for semantic segmentation.
It is a vision transformer (ViT) based deep neural network which achieved state-of-the-art
performance on the competitive Cityscapes~\cite{Cordts2016} and ADE20K~\cite{Zhou2017} segmentation
benchmarks in 2021/22.
A more detailed overview is given in appendix~\ref{sec:appendix:mask2former}.
The choice of model in \product is
largely interchangeable due to use of MMSegmentation, but Mask2Former was chosen primarily because
it performed best in the aforementioned benchmarks, and using a strong baseline for transfer
training on specific features was crucial to enable generalizability of \product.

\subsubsection{Experimental setup}\label{subsubsec:experimental-setup}
The high performance computing (HPC) platform of Imperial College London was utilized for model
training.
Models were trained using one NVIDIA RTX6000 GPU (Turing, driver 535.54.03) with 24 GB VRAM, 32 GB
of system RAM, and 4 CPU cores.

\begin{figure}[!th]
    \centering
    \begin{subfigure}[b]{0.49\textwidth}
        \includegraphics[width=\textwidth]{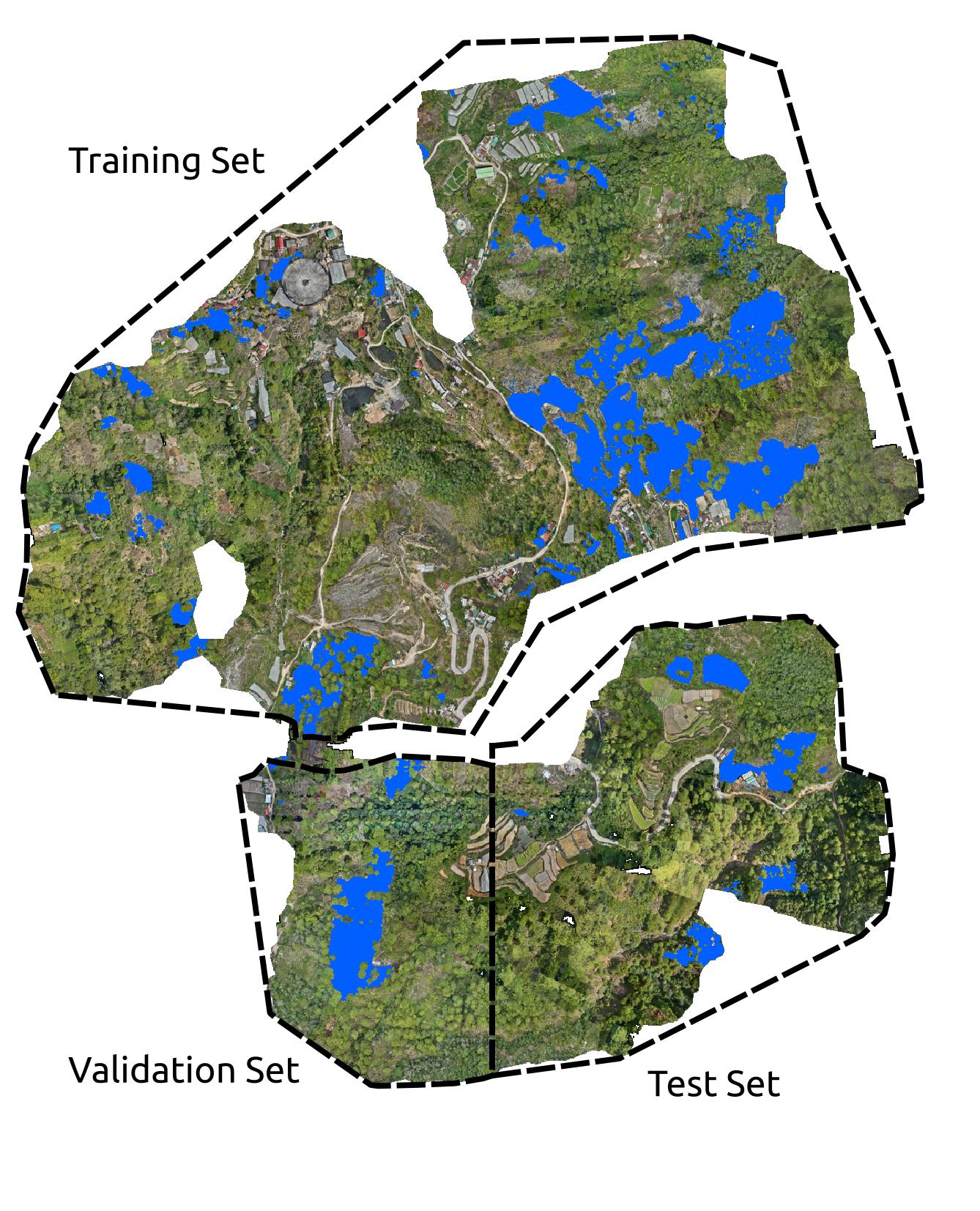}
        \caption{Dataset split for Chayote.}
        \label{fig:subfig_a}
    \end{subfigure}
    \begin{subfigure}[b]{0.49\textwidth}
        \includegraphics[width=\textwidth]{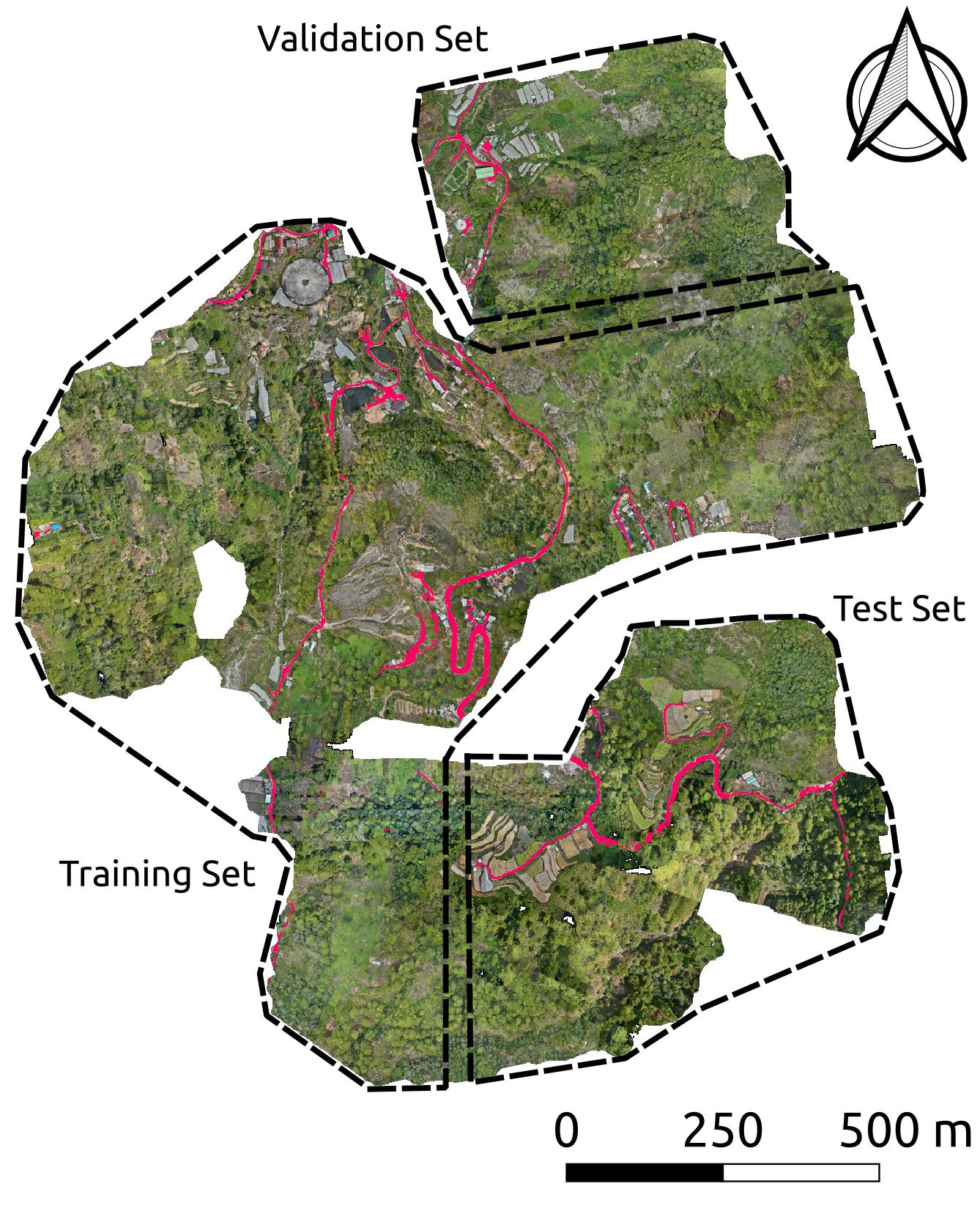}
        \caption{Dataset split for roads.}
        \label{fig:subfig_b}
    \end{subfigure}
    \caption{
        Training, validation, and test sets after manually labeling and splitting the
        Sto.\ Niño datset.
        Ground truth labels are indicated, showing a favorable distribution of classes in all three
        sets for both features.
        Spacing was introduced between the train and evaluation sets to reduce spatial leakage.}
    \label{fig:dataset-split}
\end{figure}

All runs utilized the same CUDA (11.7), PyTorch (2.0.1+cu117) and TorchVision (0.15.2+cu117)
versions, which are fixed by EcoMappers`s \verb|pip| environment.
The same seed for all libraries was used throughout the research project.
These precautions alone do not guarantee identical results on
other hardware, but they make up for the largest fluctuations in model performance.
Fixing the batch size and seed guarantees that the model receives samples in the same
order on each training run, regardless of the hardware used.

To mitigate spatial leakage and achieve a stable class distribution, the orthomosaic and label maps
were split manually into approximately 70\% train, 10\% validation, and 20\% test
data, as illustrated in Fig.~\ref{fig:dataset-split}.
Models were configured with a Swin-S backbone~\cite{Liu2021} pretrained on the Cityscapes
dataset~\cite{Cordts2016}; given the substantial differences in scenery and viewing angle in
the Sto.\ Niño dataset, the backbone was not frozen to give the model more
flexibility during transfer training.

For each resolution and degradation method, models were transfer-trained over 90 epochs with a batch
size of 12 and learning rate of $1\mathrm{e}{-4}$, using the AdamW optimizer~\cite{Loshchilov2019}.
With the default Mask2Former configuration in MMSegmentation, the optimizer used a weight
decay of 0.05, $\epsilon=1\mathrm{e}{-8}$, and $\beta=(0.9, 0.999)$.
Except for two models, the validation loss during training always plateaued before
reaching 90 epochs.
Model weights with the lowest validation loss were chosen for testing.
No hyperparameter tuning was conducted; as detailed in section~\ref{sec:introduction}, \product aims
to apply to arbitrary features and datasets, without requiring adjustments on a per-use-case basis.

The intersection over union (IoU) and Dice coefficient (F1 score)
were used for model evaluation, which are among the most prominent benchmarking metrics in the literature.
IoU measures how well predictions overlap with ground truth labels.
The Dice coefficient is the harmonic mean of precision and recall:
Precision describes the fraction of positive predictions that are correct, recall gives the
proportion of positive samples that the model correctly predicted as positive.
The formulas are given below.
\begin{equation}
    \text{IoU} = \frac{TP}{TP + FN + FP} = \frac{|\hat{Y} \cap Y|}{|\hat{Y} \cup Y|}\label{eq:iou}
\end{equation}

\begin{equation}
    \text{Dice coefficient} = \frac{2TP}{2TP + FN + FP} = \frac{2|\hat{Y} \cap Y|}{|\hat{Y} \cap Y|+|\hat{Y} \cup Y|} = \frac{2|\hat{Y} \cap Y|}{|\hat{Y}|+|Y|}\label{eq:f1}
\end{equation}

Where $TP$, $FP$ and $FN$ are true positives, false positives, and false negatives, respectively.
$\hat{Y}$ and $Y$ are predictions and ground truth, with $\cap$ and $\cup$ denoting intersection and union.

\section{Evaluation}\label{sec:evaluation}

\subsection{Profiling of EcoMapper performance}\label{subsec:pipeline-performance}
\begin{figure}[H]
    \centering
    \begin{subfigure}[b]{0.32\textwidth}
        \includegraphics[width=\textwidth]{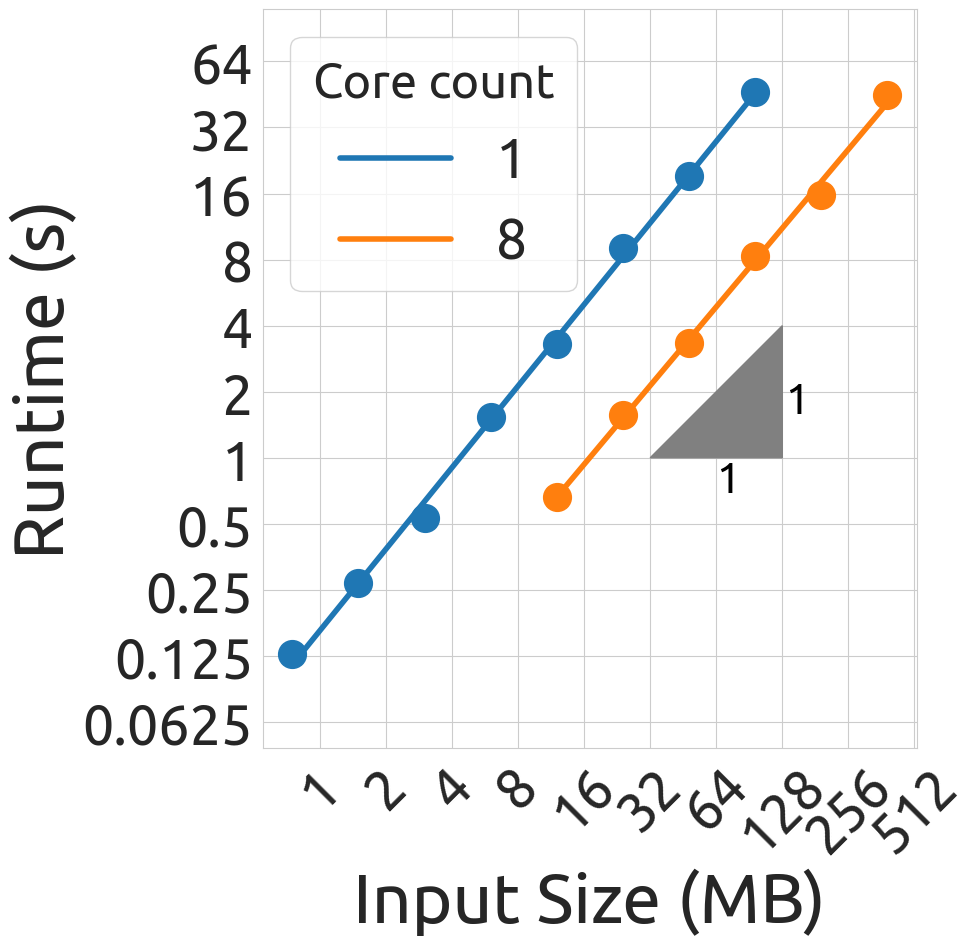}
        \caption{}
        \label{subfig:split-profiling}
    \end{subfigure}
    \begin{subfigure}[b]{0.32\textwidth}
        \includegraphics[width=\textwidth]{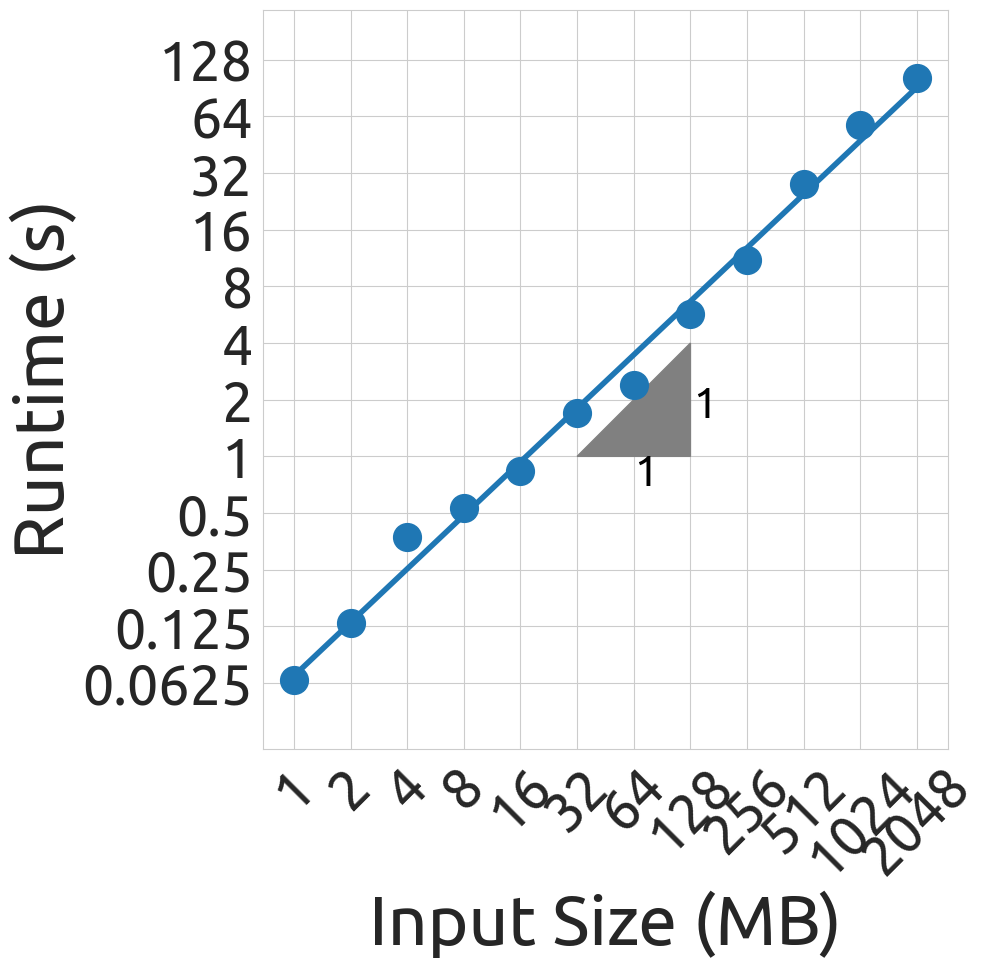}
        \caption{}
        \label{subfig:merge-profiling}
    \end{subfigure}
    \begin{subfigure}[b]{0.32\textwidth}
        \includegraphics[width=\textwidth]{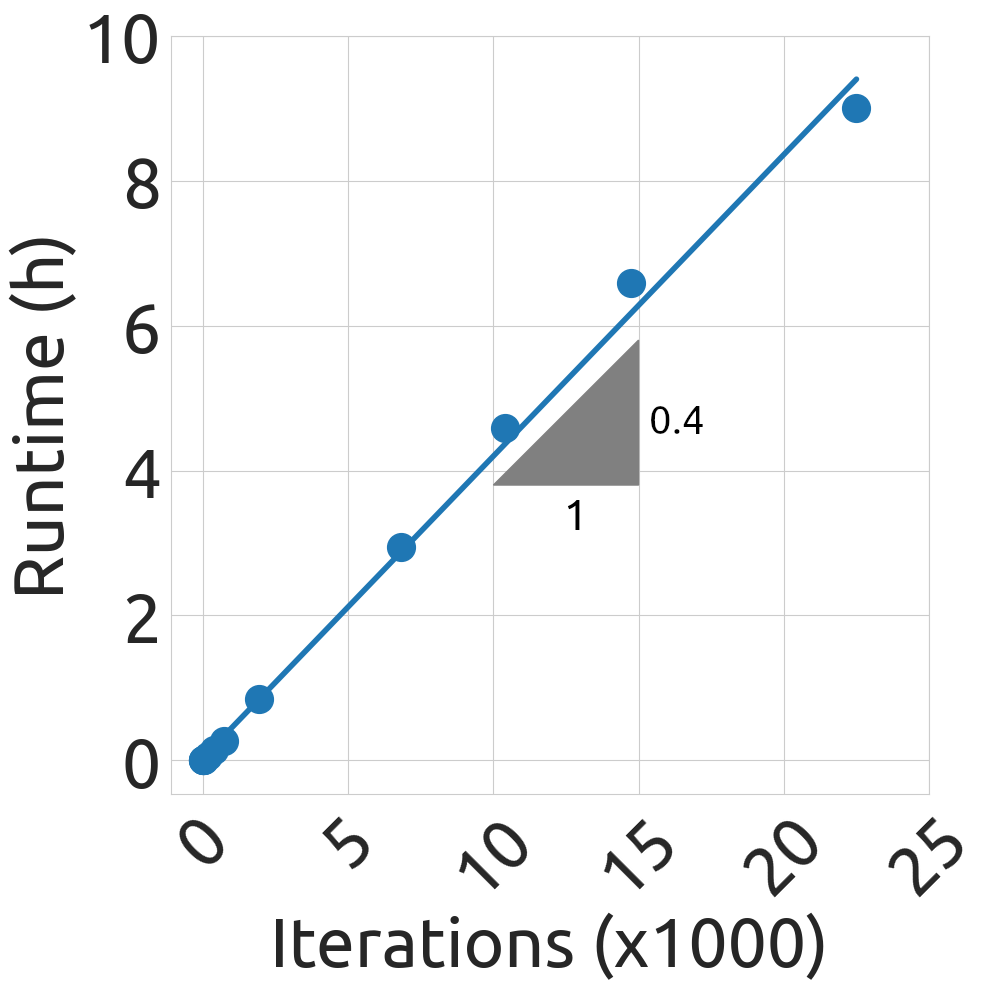}
        \caption{}
    \end{subfigure}
    \caption{\product performance in different tasks. (a) Image splitting using a stride of 0.5;
        (b) prediction merging, input size indicates total size of all tiles; (c) GPU training with
        PyTorch via MMSegmentation.
    }
    \label{fig:profiling}
\end{figure}
Code profiling was conducted to measure the efficiency of data processing and model training code of
\product, see Fig.~\ref{fig:profiling}.
Algorithms for image splitting and merging predictions naturally had time complexities of
$\mathcal{O}(nm)$, as they iterate over the $n\times m$ input image once.
Model training implemented in PyTorch also scaled linearly with input size, and benefited heavily
from GPU acceleration.

However, significant differences in runtime still arose and were managed.
EcoMapper`s algorithm to
split an input image operated at a stride of $0.5$, thus the actual amount of work per image increased by factor 4.
The operation was accelerated by distributing the image rows over all cores.
Merging tiles, while also $\mathcal{O}(n)$, was accelerated
significantly using the crop-merge algorithm, mainly due to use of vectorized NumPy operations implemented
in C to copy tile contents into the merged segmentation map.

\subsection{Semantic segmentation of Chayote and roads using \product}\label{subsec:prediction-results}
\begin{figure}[t]
    \centering
    \begin{subfigure}[b]{0.49\textwidth}
        \includegraphics[width=\textwidth]{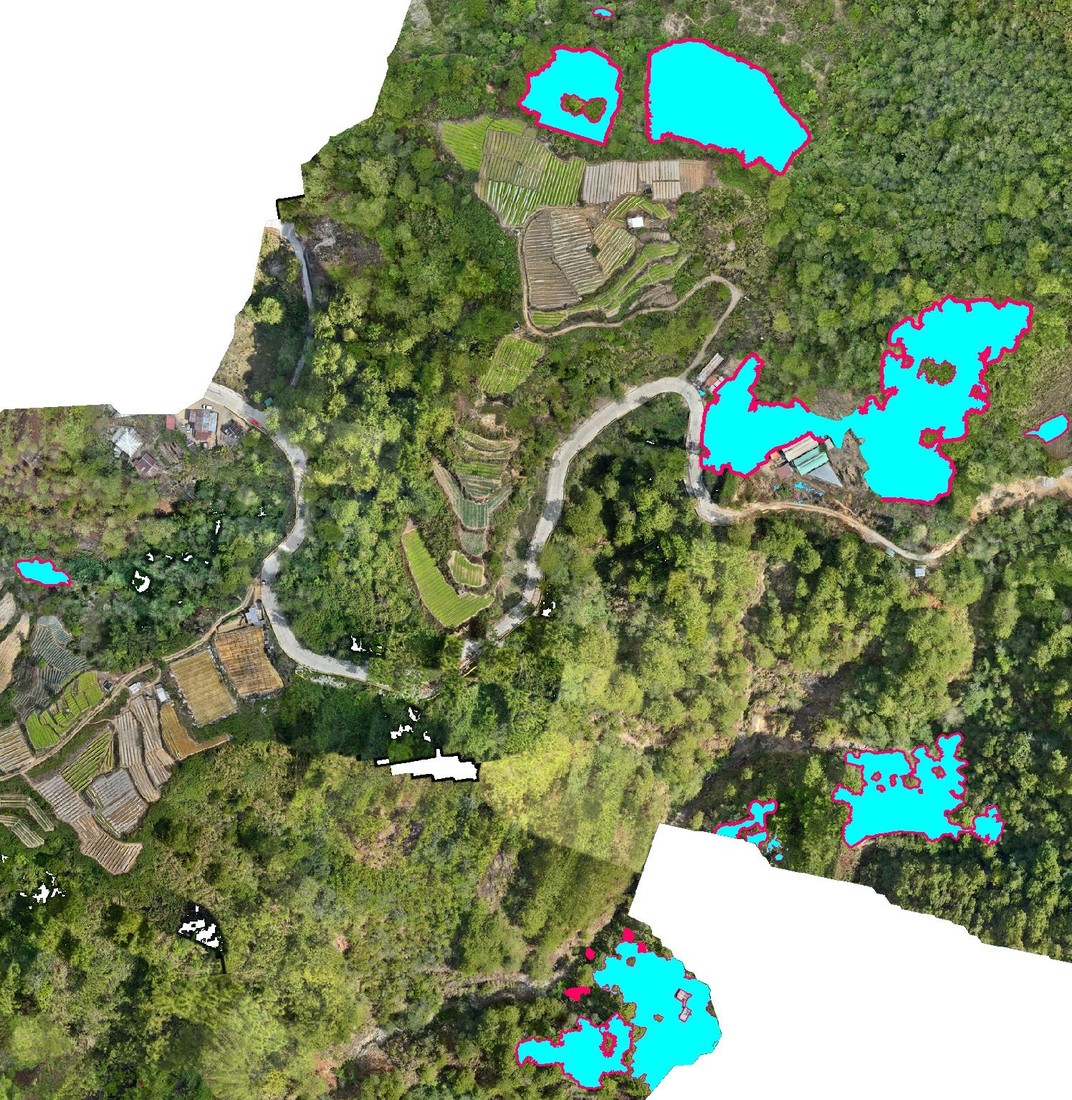}
        \caption{Chayote ground truth labels.}
    \end{subfigure}
    \begin{subfigure}[b]{0.49\textwidth}
        \includegraphics[width=\textwidth]{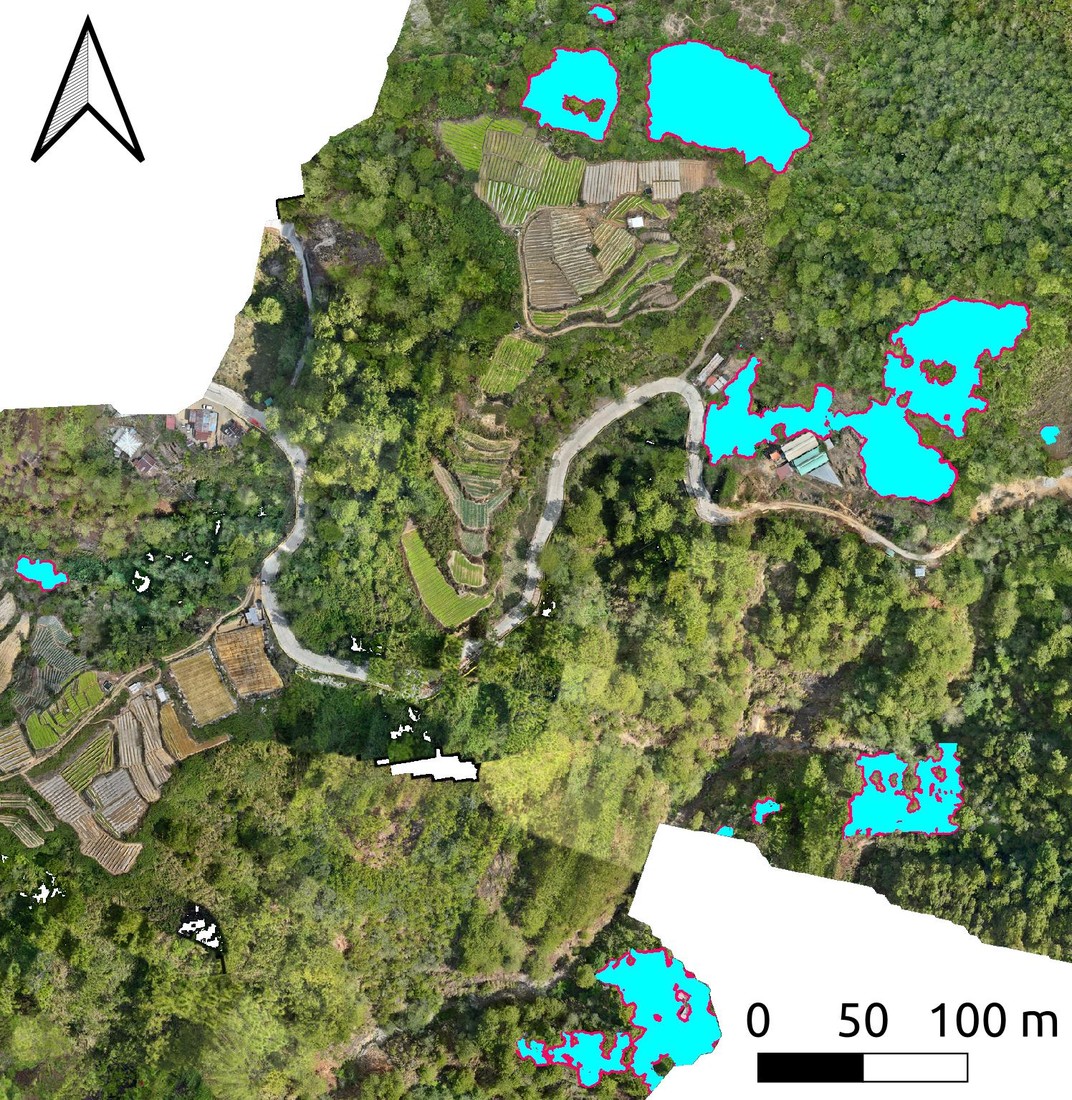}
        \caption{Predicted Chayote labels.}
        \label{subfig:chayote-pred}
    \end{subfigure}

    \begin{subfigure}[b]{0.49\textwidth}
        \includegraphics[width=\textwidth]{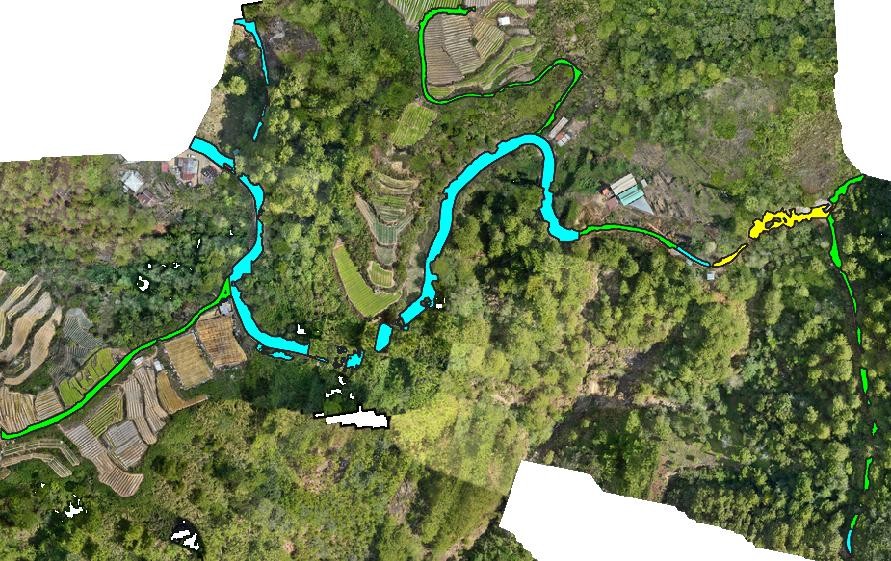}
        \caption{Road ground truth labels.}
    \end{subfigure}
    \begin{subfigure}[b]{0.49\textwidth}
        \includegraphics[width=\textwidth]{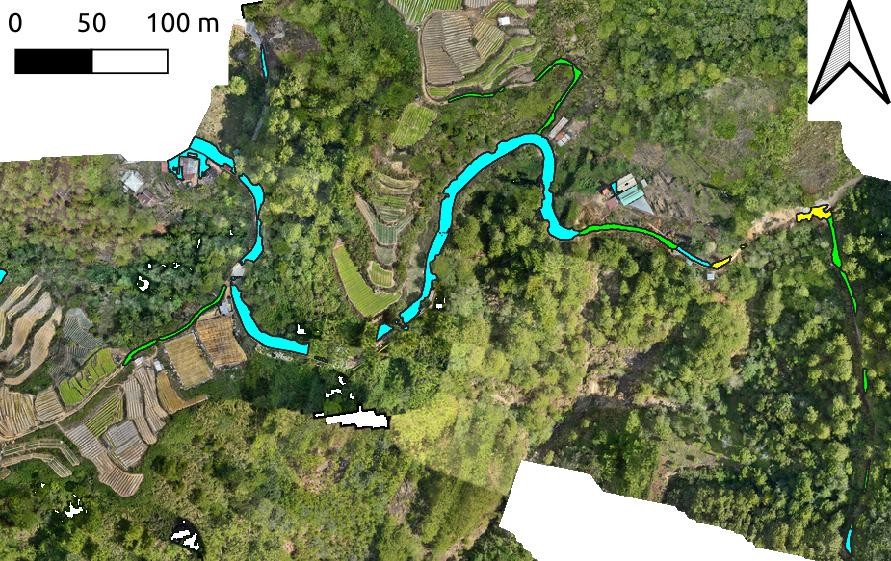}
        \caption{Predicted road labels.}
    \end{subfigure}
    \caption{Comparison of ground truth with model predictions in the test set.
    For (a) and (b), cyan indicates Chayote, red indicates border.
    For (c) and (d), cyan, green and yellow visualize asphalt, dirt, and sand roads,
        respectively (road types for illustration, not distiguished by the model).
    }
    \label{fig:test-gt-vs-pred}
\end{figure}
\begin{table}[ht]
    \caption{Per-feature test scores of Mask2Former instances trained with \product on the Sto.\
        Niño dataset (0.08 m/px). Main scores were obtained by summing the scores of individual tiles before computing the overall ratio. Scores in
        parentheses are calculated after merging perdictions into a single image using the
        crop-merge strategy.}
    \centering
    \adjustbox{max width=\textwidth}{%
    \begin{tabular}{c c c c c c}
        \hline\hline
        \rule{0pt}{3ex}
        Feature & mIoU          & mDice         & Background IoU & Feature IoU   & Border IoU \\ [1ex]
        \hline
        \rule{0pt}{3ex}
        Chayote & 0.668 (0.682) & 0.748 (0.760) & 0.986 (0.992)  & 0.792 (0.814) & 0.225 (0.240) \\
        Roads   & 0.798 (0.831) & 0.875 (0.900) & 0.995 (0.995)  & 0.610 (0.667) & n/a \\ [1ex]
        \hline
    
    \end{tabular}}
    \label{tab:pred-metrics}
\end{table}

\noindent The models trained with \product achieved mean IoUs (mIoU) of 0.668 and 0.798 for
Chayote and roads, respectively (Tab.~\ref{tab:pred-metrics}).
Chayote scores were skewed by the border class, which the model did not segment accurately.
However, labeling of this class entailed a high degree of subjectivity.
Despite the poor IoU the model still provided reasonable borders around its Chayote predictions (Fig.~\ref{subfig:chayote-pred}).
The mIoU and mean Dice coefficient (mDice) for Chayote excluding the border class were 0.889
and 0.939, respectively.

Merging predictions using the tile-crop strategy improved segmentation
performance notably.
Logit-merge scores (not shown here) were consistently lower than baseline scores.
This proves the significance of post-processing geospatial predictions, which \product also
facilitates.

\begin{figure}[ht]
    \centering
    \includegraphics[width=\textwidth]{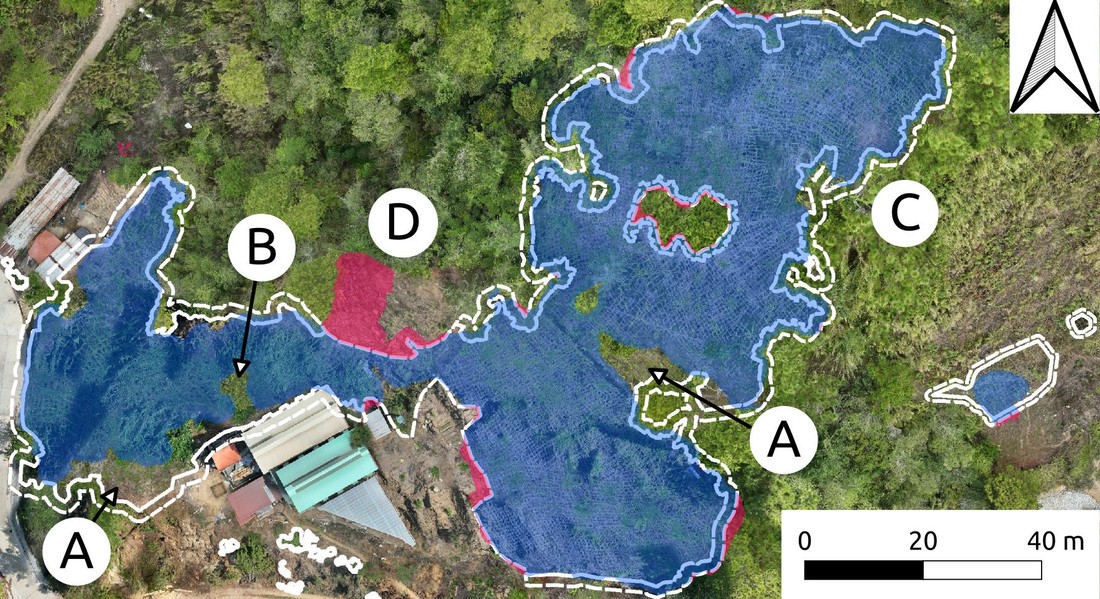}
    \caption{
        Closeup view of the difference between Chayote ground truth labels and
        predictions.
        Blue indicates agreement, red areas are false positives.
        White outline describes ground truth.
        (A) Largest disagreements occur in regions not containing Chayote, but contextually
        considered part of the plantation;
        (B) a species erronously included in the ground truth is excluded by the model;
        (C) a large amount of disagreement is incurred at the boundary, as it is unclear during
        labeling where a plantation should end;
        (D) the largest false positive in the test dataset -- it is still part of the main plantation
        and thus represents a sensible prediction, similar to the areas at (A).
    }
    \label{fig:chayote-test-gt-vs-pred-closeup}
\end{figure}

Overall prediction quality was high:
the generated segmentation maps were free of noise and incurred very few false
positives.
Moreover, models identified the main bodies of all
feature clusters, as shown in Fig.~\ref{fig:test-gt-vs-pred}.
A closeup analysis of the difference between ground truth and prediction for Chayote is given in
Fig.~\ref{fig:chayote-test-gt-vs-pred-closeup}.

\subsection{Effects of survey characteristics on per-feature model performance}\label{subsec:model-performance}
\begin{figure}[H]
    \centering
    \begin{subfigure}[b]{0.475\textwidth}
        \includegraphics[width=\textwidth]{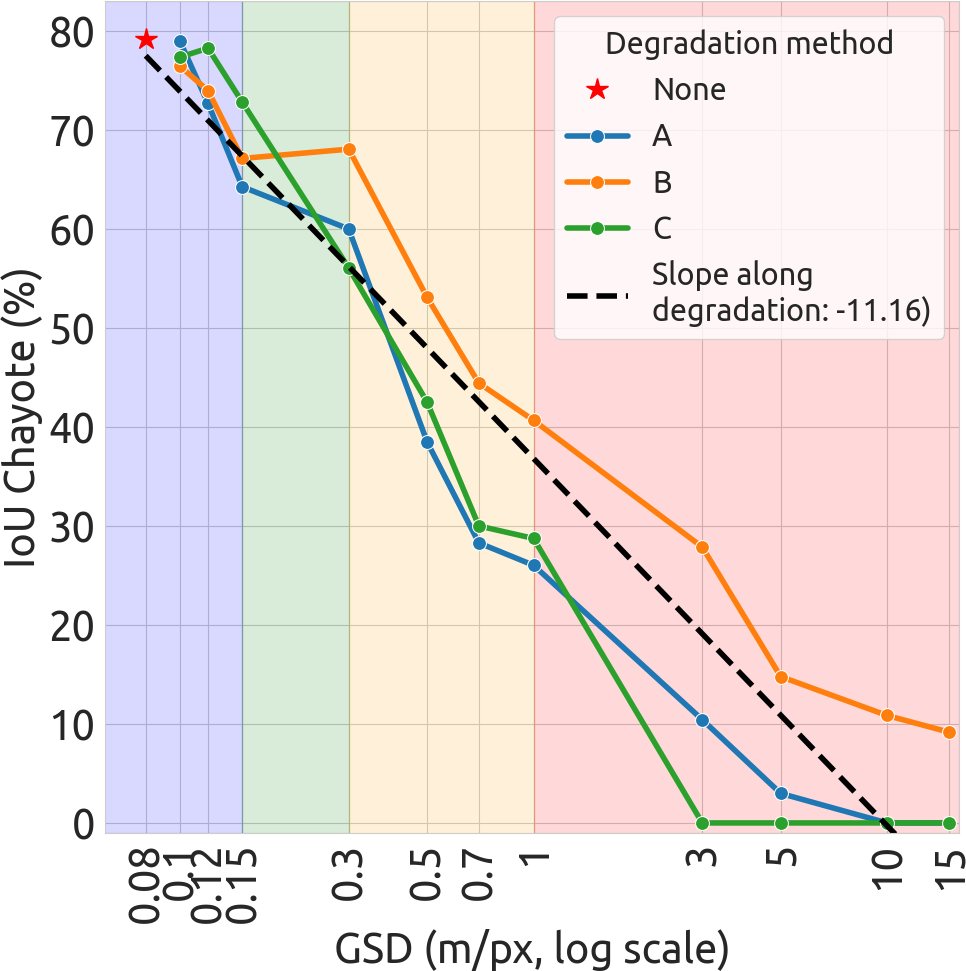}
        \caption{}
        \label{subfig:degradation-plot-chayote}
    \end{subfigure}
    \begin{subfigure}[b]{0.475\textwidth}
        \includegraphics[width=\textwidth]{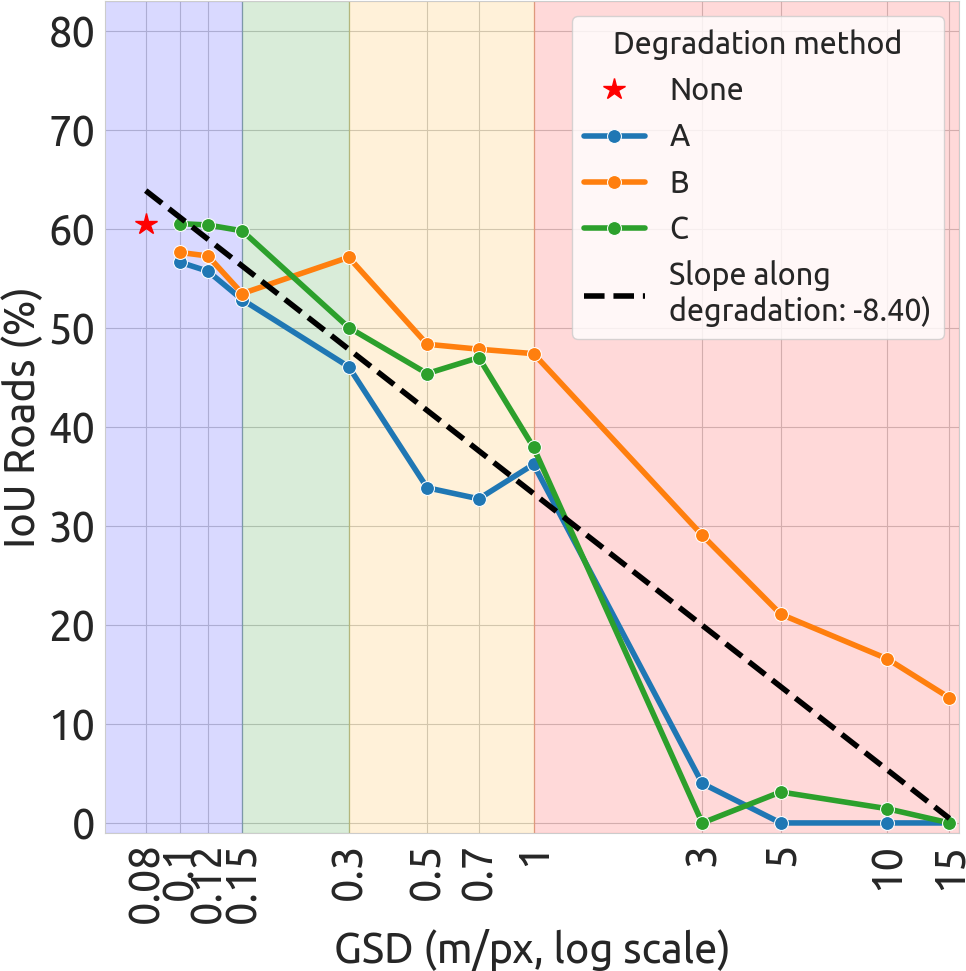}
        \caption{}
        \label{subfig:degradation-plot-roads}
    \end{subfigure}

    \begin{subfigure}[b]{0.475\textwidth}
        \includegraphics[width=\textwidth]{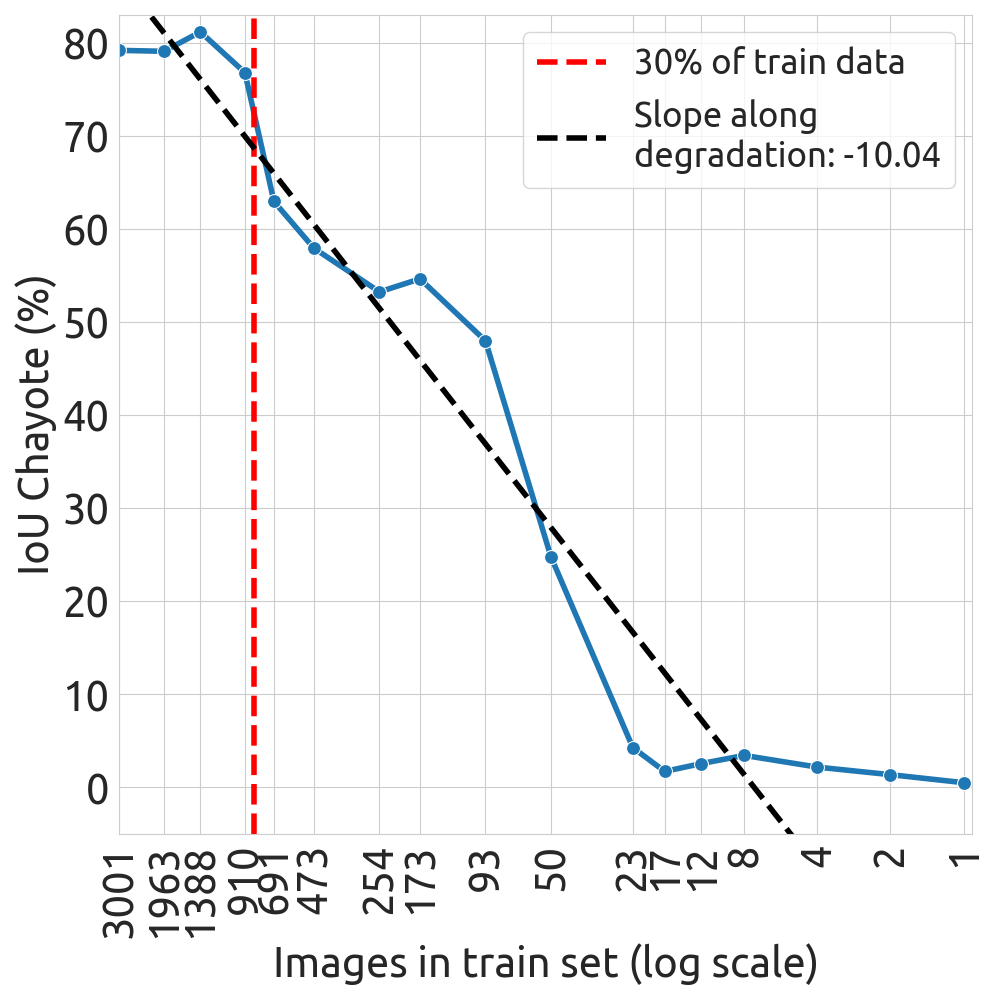}
        \caption{}
        \label{subfig:chayote_train_extent}
    \end{subfigure}
    \begin{subfigure}[b]{0.475\textwidth}
        \includegraphics[width=\textwidth]{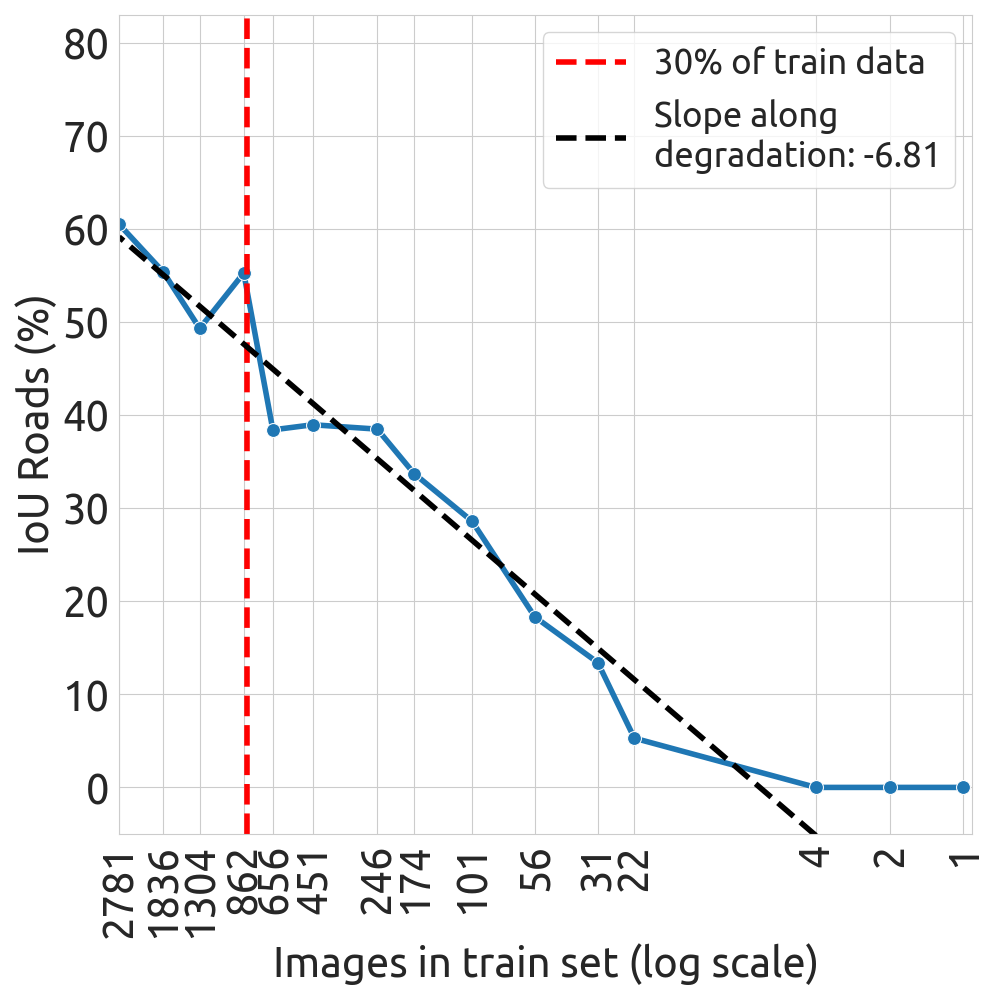}
        \caption{}
        \label{subfig:roads_train_extent}
    \end{subfigure}

    \caption{Chayote (a, c) and road (b, d) IoU model scores relative to GSD and dataset
    size. Slight differences in image counts in (c) and (d) are due to the difference in splits
    shown in Fig.~\ref{fig:dataset-split}.
    Background colors in (a) and (b) indicate which RS products provide the shown GSDs.
    Blue: UAVs; green: Maxar satellites with super-resolution post-processing; yellow: VHR commerical satellites; red: non-commercial satellites.}
    \label{fig:degradation-plot}
\end{figure}

Fig.~\ref{fig:degradation-plot} visualizes model performance w.r.t.\ principal survey
characteristics (GSD, feature size, dataset size).
Corresponding tables are given in appendix~\ref{sec:appendix:degradation-scores}.
An exemplary visualization of GSD impact on Chayote segmentation performance is given in
appendix~\ref{sec:appendix:degradation-visualization}.

While the Chayote model performed better at high resolutions, the performance dropped faster
compared to the road model as GSD increased.
As roads are generally easier to detect at coarser resolutions, the road model was robuster to
changes in GSD\@, as illustrated by the slopes of the best fit lines.
Beyond 0.5 m/px, the IoU for Chayote fell below the IoU for roads.

A sharp drop in Chayote performance was observed initially at 0.12 m/px.
For roads, the performance was stable up to 0.12 m/px and then started to decrease slowly, but
stabilized between 0.5 and 1 m/px before dropping sharply.
As also observed in~\cite{Barbedo2019}, segmentation performance sometimes
improved when reducing GSD\@, which the results in Fig.~\ref{fig:degradation-plot} confirm.

Mean segmentation performance across degradation methods at 0.15 m/px was
within 85.91\% and 91.48\% of the highest score for Chayote and roads, respectively.
Findings thus suggest that VHR satellite products, such as super-resolution Maxar imagery, may
provide a viable alternative to UAV imagery for VHR feature extraction.
\noindent Dataset size also had a severe impact on model performance, visualized by the
slopes of best fit lines.
Model performance started to decline strongly when training on less than 30\% of tiles from the
original training set (indicated by red lines).
The slope in Fig.~\ref{subfig:chayote_train_extent} was again steeper compared to
Fig.~\ref{subfig:roads_train_extent}, suggesting that more images are required to train a model on
smaller features.
The intermediate performance plateau in Fig.~\ref{subfig:roads_train_extent} is comparable to the
plateau in Fig.~\ref{subfig:degradation-plot-roads}, and suggests that loss preceding the plateau
stems from segmentation of smaller dirt roads.
Larger asphalted roads are then segmented reliably until much stronger dataset degradations
occur.

\section{Discussion}\label{sec:discussion}

\subsection{\product did not require dataset-specific tuning to achieve strong segmentation performance}\label{subsec:discussion-ecomapper-native-segmentation-competetive}
\product mIoU and mDice scores for Chayote and roads were comparable to other UAV-based studies using ViT
networks; c.f.~\cite{Zhou2022, Gomes2022, Gibril2023, Huang2023} for studies of features comparable
to Chayote, and~\cite{Wang2022, Zhang2022, Kumar2022, Gu2022} for studies on roads and similar
features.

All cited works designed their networks based on the dataset at hand.
\product models were not tuned but their segmentation performance matched performance figures of prior studies.
The first research objective to design a framework for arbitrary segmentation was thus
achieved, without compromising on segmentation performance.
Results show that transfer training of ViT networks represents a compelling alternative to
redesigning models on a per-study basis.

\subsection{Relationship between GSD and feature size}\label{subsec:cording-index}
The results in Fig.~\ref{fig:degradation-plot} show that several GSDs allowed for accurate
feature segmentation.
It is desirable to identify the maximum GSD before performance deteriorates, to ensure
that data collected in surveys are suitable for analysis with DL\@.

The Cording Index (\metric) proposed below provides a lower and upper bound on the ``critical GSD''
at which model performance declines.
Surveys may orient their choice of GSD based on this interval.
The index was established from empirical study of the results in section~\ref{sec:evaluation}, and
from figures reported in prior works~\cite{Brown2022, Ilniyaz2022}.

\begin{enumerate}
    \item Let $F$ be the feature to identify.
    \item Determine the smallest visible attribute (SVA) of $F$ in overhead imagery
    that contributes to the uniqueness or identifiability of $F$.
    \item Repeatedly measure the SVA size in various locations across the dataset.
    For rectangular shapes, measure the shorter side length.
    For circular shapes, measure the diameter.
    Highly irregular shapes cannot be measured directly; an average of the lengths of
    segments may be chosen, but first ensure the SVA was selected correctly.
    \item Using the measurements, determine the upper and lower bounds of the SVA's size, $f_{s_1}$ and $f_{s_2}$.
    \item The critical GSD for $F$ is then suggested to lie in the following interval:
\end{enumerate}

\begin{equation}
    \text{GSD}_F \in (\frac{f_{s_1}}{3}, \frac{f_{s_2}}{3})\label{eq:the-metric}
\end{equation}
\metric aims to balance model performance and surveying duration.
GSDs far below $\frac{f_{s_1}}{3}$ will thereafter not result in large performance gains.
Conversely, GSDs far above $\frac{f_{s_2}}{3}$ will not offer the model sufficient information to
identify $F$ without incurring significant error.

Below it is shown that \metric can be applied to obtain sensible GSD intervals for several
features and models in this study and prior research.

\subsubsection{GSD prediction for Chayote, roads, and other features}
The SVA for Chayote are its leaves, which are heart-shaped and in the dataset had a diameter of
15--35 cm.
Using \metric, the critical GSD in m/px for this feature is in
$(\frac{0.15}{3}=0.05,\frac{0.35}{3}\approx0.117)$.
This interval is appropriate, as Chayote segmentation performance declined steadily beyond
0.10--0.12 m/px and improved marginally towards 0.08 m/px (c.f.\ Fig.~\ref{fig:degradation-plot}).

Asphalted roads in the dataset lacked markings and had no other SVA\@.
Their width ranged from 3--8 m.
The SVA of smaller dirt roads were tire tracks, which distinguished them from farmland.
Individually, these tracks were found to be approximately 0.4--0.85 m wide.

\metric then gives $(\frac{0.4}{3}\approx0.13,\frac{0.85}{3}\approx0.28)$ as GSD interval
in m/px for dirt roads.
Reviewing Fig.~\ref{fig:degradation-plot}, the performance was initially stable, confirming that
GSDs below the lower bound did not significantly improve the model's performance.
In addition, performance started to deteriorate around 0.12--0.15 m/px, with a strong dip for all
degradation methods from 0.3 m/px to 0.5 m/px.
This indicates that most dirt roads in the dataset were very difficult to identify above 0.3 m/px,
coinciding closely with the upper bound of \metric.

Road segmentation performance stabilized from 0.3 to around 1 m/px, suggesting that large
asphalted roads being were easily detectable in this GSD range.
For asphalted roads, the GSD interval according to \metric is $(\frac{3}{3}=1,\frac{8}{3}\approx2.67)$.
This explains the drop in performance from 1 to 3 m/px, and the stable performance from 0.3--1 m/px.

In appendix~\ref{sec:appendix:ca-examples}, additional applications of \metric to other features
from previous studies are made.
\metric provides sensible GSD intervals for these cases, even though different models and datasets
were used.

\subsection{Tradeoffs between survey characteristics and DL performance}\label{subsec:discussion-degradation-methods}
\metric can provide an optimal GSD range for a specific feature, to accelerate data collection
without deteriorating model performance.
However, sampling at maximum quality can make data more versatile for future analysis of small-scale
features.
A tradeoff between surveying resources, features to study, and model performance must therefore
be made.

Surveying extent, simulated with dataset size, must also be considered.
Based on Fig.~\ref{fig:degradation-plot}
at least 30\% of samples from the initial training set were needed to achieve adequate segmentation
performance.
Performance gains from using additional data were unclear.

Models may perform equally well with fewer training samples on a test set of the same study
region.
However, DL models generalize better when trained on more data points~\cite{Garcia2018, Soekhoe2016}.
If a model should be reapplied to other regions in the future, training on a larger dataset may
prove beneficial.
The scope of model application thus needs to be taken into account when choosing survey extent.

\subsection{Proposal for an integrated DL-RS field survey workflow}
\begin{figure}[ht]
    \centering
    \includegraphics[width=0.94\textwidth]{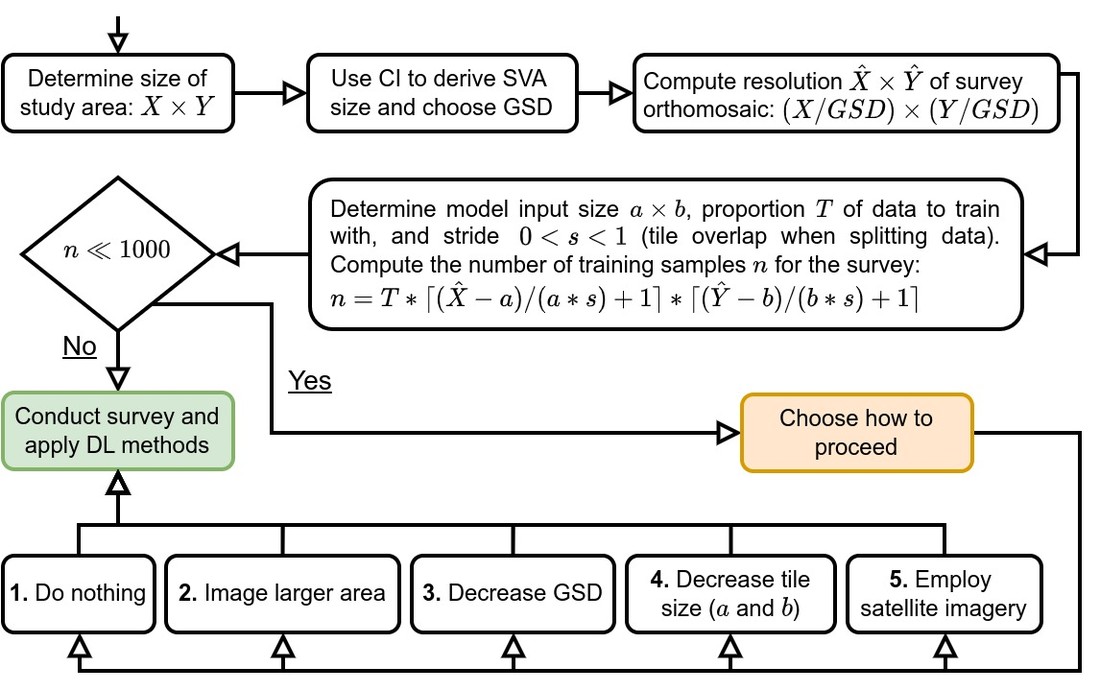}
    \caption{Field survey methodology with mechnasims to accomodate for DL model requirements.
    Survey feasibility is informed through study extent, feature size, GSD, and the
    number of training samples resulting from these characteristics.
    Actions are proposed for cases of data shortage, see text for details.
    }
    \label{fig:survey-method}
\end{figure}
A surveying workflow was developed from the studied effects of FSCs on DL performance.
The workflow is illustrated in Fig.~\ref{fig:survey-method}.
Feature size, survey area, and GSD estimated by \metric were integrated to balance the tradeoff
between image resolution, survey resources, and DL performance.

\noindent Critically, additional survey adjustments may be needed to ensure
sufficient data is collected.
In case of data shortage, five possible actions with different implications on
survey characteristics and model performance are proposed:

\begin{enumerate}
    \item Do nothing:
    Survey duration is unaffected, but model performance may be impaired if the dataset
    is too small, due to overfitting (Figs.~\ref{subfig:chayote_train_extent} and~\ref{subfig:roads_train_extent}).
    \item Image larger area:
    Survey duration increases, but the model receives sufficient data at the desired GSD\@.
    \item Decrease GSD (by reducing imaging altitude or using a higher fidelity sensor)\@:
    Survey duration increases, the model receives enough data, but may perform suboptimally
    if unnecessary details are captured.
    Simultaneously, VHR imagery can be synthetically downscaled and allows for
    future analysis of smaller features.
    Exemplary implications of GSD on imaging extent and surveying duration are described in
    appendix~\ref{sec:appendix:pixel-density-area}.
    \item Reduce tile size:
    Survey duration is unaffected, the model receives sufficient samples, but smaller tiles may
    impact model performance.
    \item Employ satellite data: Commercial VHR satellite products may present a viable alternative
    to UAV surveying, as demonstrated by model performance figures at 0.15 m/px in
    Fig.~\ref{fig:degradation-plot}.
    However, satellite imagery can be nosier and occluded by clouds, thus performance figures from
    this UAV-based study may not directly translate to satellite use.
\end{enumerate}
Future works should investigate the implications of outlined choices on model performance, and
attempt to apply the proposed workflow together with \product for a comprehensive feature study.

\section{Conclusion}\label{sec:conclusion}
This study proposed \product, a pipeline for segmentation of arbitrary features in RS imagery.
Using \product, it was shown that two visually distinct features can be successfully segmented in a
complex, real-world drone dataset.

Code profiling proved that \product achieves high computational performance in data processing and
model training tasks.
Segmentation performance without dataset-specific tuning was competitive with results of
comparable studies which designed models for particular datasets.
The proposed Cording Index and DL-informed workflow for RS surveys provide a foundation
to understand and leverage the synergies of DL and RS\@, with the aim to accelerate field surveys
and make data collection more effective.

\product predictions enable species specific analysis over the domain, such as shown in
appendix~\ref{sec:appendix:example-of-product-applicability}, thereby supporting systems of
precision agriculture, estimation of biodiversity, and many more.
Additional work is needed to lower computational requirements and encourage
deployment of DL-based solutions on less powerful platforms.
Future studies to segment RS imagery may apply \product to avoid the need for RS data processing
and model implementation.

\vfill
\subsection*{Funding}
The authors acknowledge that they did not receive funding for this work.

\subsection*{Conflicts of Interest}
The authors declare that there is no conflict of interest regarding the publication of this article.

\newpage

\printbibliography

\appendix
\renewcommand{\theequation}{\thesection.\arabic{equation}}
\renewcommand{\thefigure}{\thesection.\arabic{figure}}
\renewcommand{\thetable}{\thesection.\arabic{table}}
\renewcommand{\thealgorithm}{\thesection.\arabic{algorithm}}
\setcounter{equation}{0}
\setcounter{figure}{0}
\setcounter{table}{0}
\setcounter{algorithm}{0}

\newpage
\section{Application of the Segment Anything Model (SAM) to a section of the Sto.\ Niño dataset}\label{sec:appendix:sam}

\begin{figure}[ht]
    \centering
    \begin{subfigure}[b]{0.49\textwidth}
        \includegraphics[width=\textwidth]{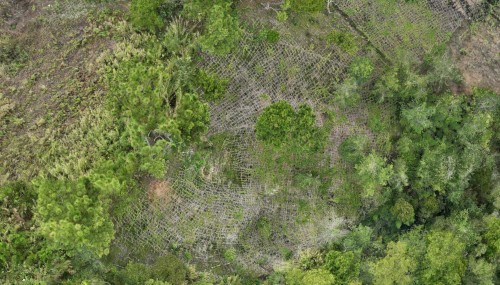}
        \caption{}
        \label{fig:sam-input}
    \end{subfigure}
    \begin{subfigure}[b]{0.49\textwidth}
        \includegraphics[width=\textwidth]{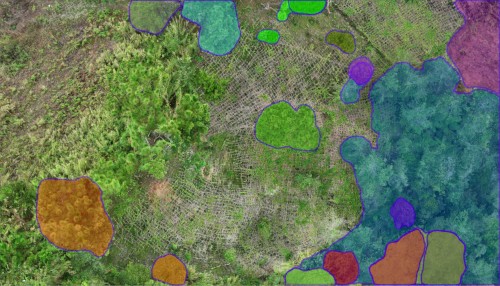}
        \caption{}
        \label{fig:sam-auto}
    \end{subfigure}

    \begin{subfigure}[b]{0.49\textwidth}
        \includegraphics[width=\textwidth]{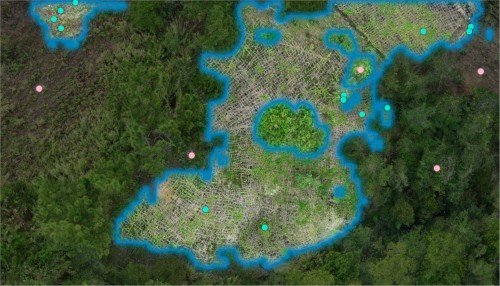}
        \caption{}
        \label{fig:sam-guided}
    \end{subfigure}
    \begin{subfigure}[b]{0.49\textwidth}
        \includegraphics[width=\textwidth]{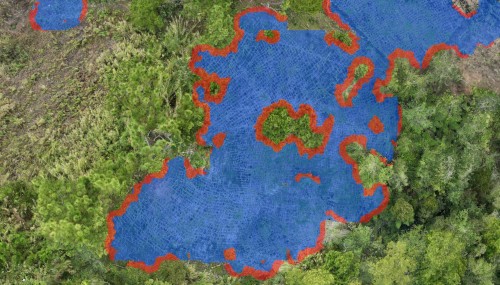}
        \caption{}
        \label{fig:sam-ecomapper}
    \end{subfigure}
    \caption{
        Application of SAM to a Chayote plantation from the test portion of the Sto.\ Niño dataset.
        (a) Input image; (b) SAM segmentation in automatic mode; (c) SAM segmentation after 18
        adjustment clicks by the user; (d) automatic segmentation by the model trained with
        \product (blue indicates Chayote, red border indicates model uncertainty).}
        \label{fig:sam}
\end{figure}

The Segment Anything model (SAM) is part of Meta's Segment Anything
project~\cite{kirillov2023segany}, which consists of several novel segmentation tasks, a new dataset
for segmentation containing 11M images and 1.1B masks, and SAM itself.
SAM is a zero-shot network which extracts segmentation masks from images under a given task at
interactive speed.
Meta has made the model available for several tasks in an online demo, which allows users to segment
objects over a large collection of images.
A section of the Sto.\ Niño dataset was uploaded to this demo, showing a Chayote
plantation and its surroundings, and the two main segmentation tasks of SAM were run on this image.
Fig.~\ref{fig:sam} illustrates the experiment, and compares the result with the predictions of a
model trained by \product.
Note that the image is part of the test set, i.e., it was not shown to the model during training.
\newpage

\section{Overview of the Sto.\ Niño dataset}\label{sec:appendix:dataset}
\begin{figure}[ht]
    \centering
    \includegraphics[scale=0.2]{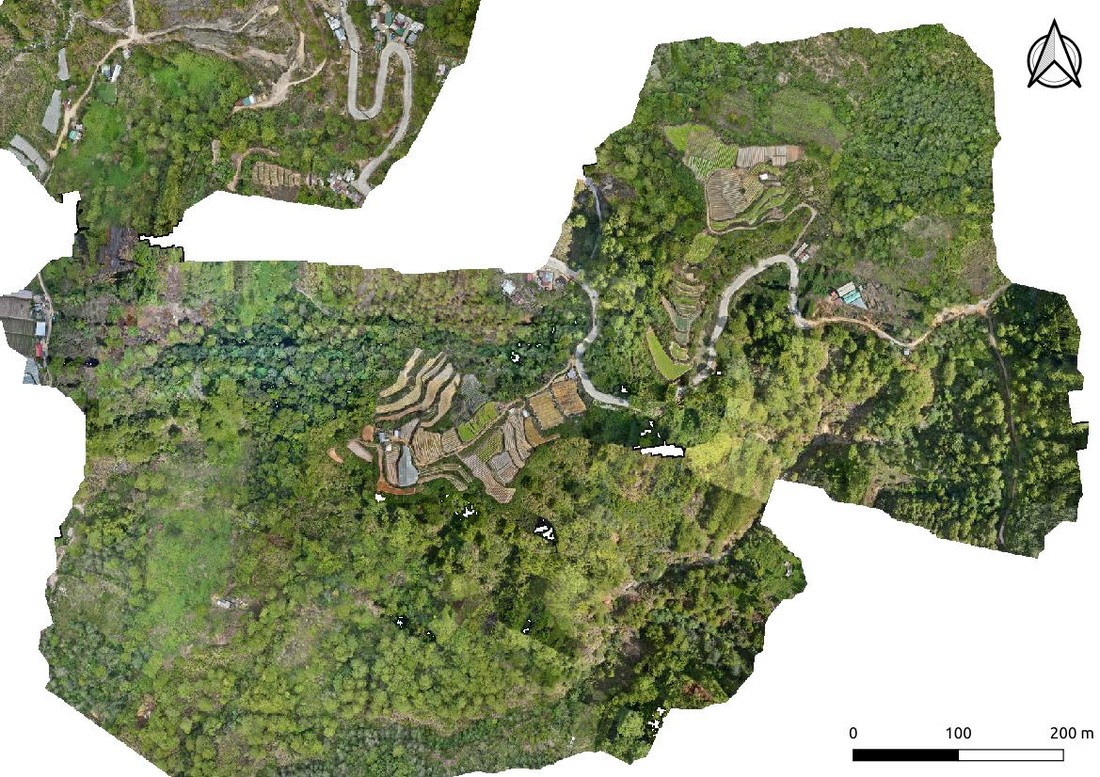}
    \caption{
        An excerpt of the Sto.\ Niño dataset, showing large varieties of vegetation, infrastructure,
        and image quality. The native resolution of around 2cm/px remains fixed over the entire
        dataset.
        }\label{fig:dataset}
\end{figure}

\section{Labeling tutorials for QGIS and CVAT}\label{sec:appendix:labeling-tutorials}
The first tutorial series\footnote{\url{https://www.youtube.com/playlist?list=PLpVXG8nY0_inhvIa8aV7kWRqr4RjWe5BU}}
details the labeling process in QGIS\@.
It first walks through importing the dataset and creating image pyramids for accelerated navigation
within the software.
Then, the various labeling tools are showcased.
Finally, the export process is detailed to generate segmentation maps in GeoTiff format.

Readers can also refer to an alternative tutorial\footnote{\url{https://youtu.be/LK6WaPUBmO8}} for using the CVAT
labeling tool, which partially automates label creation.
This video explains how to first generate image tiles with \product, upload them to CVAT
for labeling, and finally import the generated labels back into \product.

\section{View of a Chayote plantation in the Sto.\ Niño dataset}\label{sec:appendix:chayote-plantation-example}
\begin{figure}[ht]
    \centering
    \includegraphics[width=0.8\textwidth]{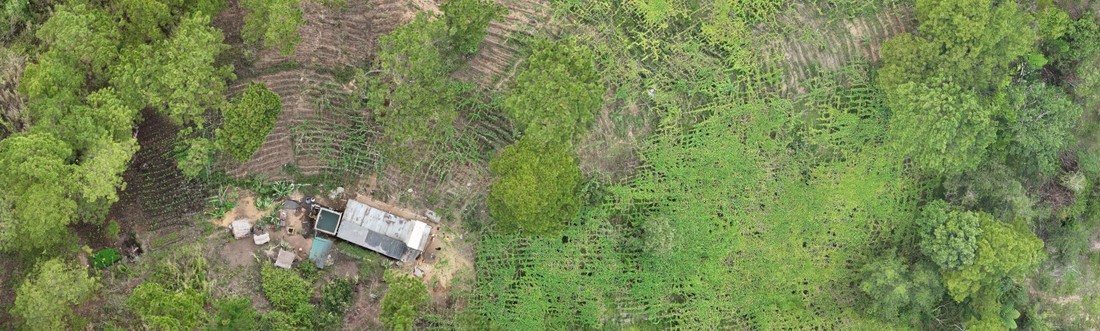}
    \caption{
        A partial view of a Chayote plantation in the Tublay region. The trellises on
        which the fruits grow form a distinct grid-like shape when viewed from above.
        }\label{fig:chayote-plantation-example}
\end{figure}

\newpage
\begin{landscape}
    \section{\product Task Network}\label{sec:appendix:product-task-network}

    \begin{figure}[ht]
        \centering
        \includegraphics[scale=0.40]{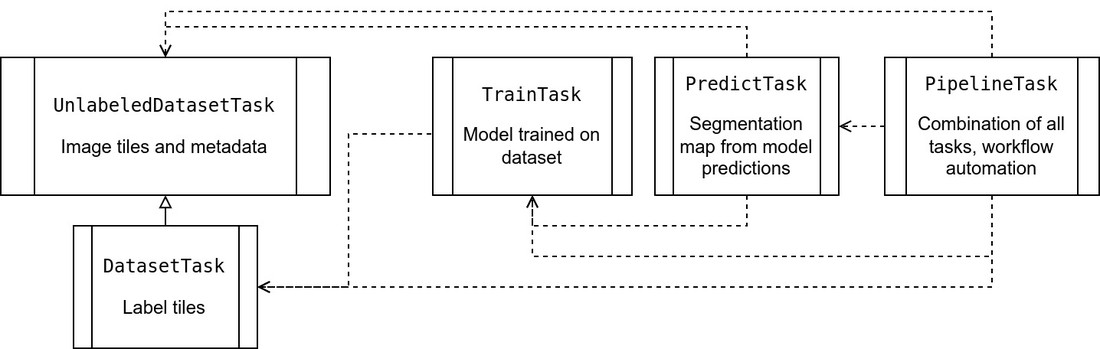}
        \caption{The \product network of Tasks, using UML notation. DatasetTask
        inherits from UnlabeledDatasetTask to introduce labels.
        Dashed arrows indicate dependencies.}
        \label{fig:task-network}
    \end{figure}
    \fillandplacepagenumber
\end{landscape}

\newpage
\section{Notable features of \product}\label{sec:appendix:notable-features-of-pipeline}

\subsection{Parallelized processing of large inputs}
Data processing in \product was parallelized using Python's
\verb|concurrent.futures| module.
The contained \verb|ProcessPoolExecutor| creates a process for each job to run, instead of using
threads.
This allows side-stepping the global interpreter lock in Python which otherwise prevents
multiple threads from executing Python byte code simultaneously.
\product also chunks input data in advance to distribute the workload effectively across all cores.
Inputs of arbitrary size are handled by using memory mapping.

\subsection{Failsafes through journaling}
Data processing and model training tasks run for long durations and can be interrupted due to power
outages or unexpected resource constraints.
\product atomically commits the progress of Tasks to a journal, which is read when restarting after
an interruption.
This allows the pipeline to continue computation from the last stable state.

\subsection{Hardware compatibility}
\product interfaces with the PyTorch training backend to allow training on a wide range of
hardware configurations.
The code has been tested on CPU-only systems, consumer-grade systems with a single GPU, as well as
HPC environments with up to 8 GPUs.
Computational requirements for training are adjustable by selecting a particular model
variant during model selection, providing trade-offs between training cost and prediction quality.

\subsection{Supported data formats and CVAT integration}
The image and label map inputs to \product can have any common format, and in particular geospatial
formats such as GeoTiff for images and GeoJSON and Shapefiles for label maps are supported.
\product also supports reading annotations generated with CVAT to help accelerate the labeling
procedure.

\newpage
\section{Strategies for processing geospatial data in a DL context}\label{sec:appendix:stragies-for-merging-tile-predictions}
\subsection{Calculating sample weights for training}\label{subsec:appendix:sample-weights}
\begin{algorithm}[ht]
    \begin{algorithmic}[1]
        \Require train\_dataset\_label\_tiles
        \State $\text{class\_occurrences} \gets \text{An empty list of lists}$

        \For{label\_tile in train\_dataset\_label\_tiles}
            \State $m \gets \text{Load label\_tile as a grayscale image}$ \Comment{Value of each pixel in $m$ is its class}
            \State $\text{Append the number of occurrences of each class in } m \text{ to class\_occurrences}$
        \EndFor

        \State $\text{total\_class\_occurrences} \gets \text{Sum of lists in \text{class\_occurrences}}$
        \newline
        \State $\text{sample\_weights} \gets \text{Empty list}$
        \For{sample\_class\_occurrences in class\_occurrences}
            \State $s \gets \text{An empty list}$
            \For{i, class\_occurrence in enumerate(sample\_class\_occurrences)}
                \State $\text{Append class\_occurrence / total\_class\_occurrences[i] to }s$

                \Comment{Proportion of class $i$ in this tile, relative to proportion of $i$ in dataset}
            \EndFor
            \State $\text{Append the sum of items in }s\text{ to sample\_weights}$
        \EndFor
        \State $\text{return sample\_weights}$
    \end{algorithmic}
    \caption{Calculation of sample weights for the train dataset. Samples containing minority
    classes are weighted more strongly, and thus have higher probability of being drawn when a batch
    of training data is created.}
    \label{alg:sample-weights}
\end{algorithm}

\subsection{Merging predictions on model logits}\label{subsec:appendix:logit-merging}
The trained model generates per-pixel predictions for each image tile it is given.
Unprocessed, such predictions are referred to as logits, which are ``confidence scores'' of the
model to assign a pixel to a given class.
Each image tile has an associated logit tile after predictions are completed.
Due to tile overlap, these predictions cannot be stitched together into a segmentation map directly.

Let $L$ denote the set of logit tiles overlapping at a pixel $p$, and $T$ the set of
image tiles corresponding to $L$.
A logit tile $l_t \in L$, corresponding to image tile $t \in T$, has shape $(h, w, |C|)$, where
$h, w$ are the tile dimensions, and $C$ the set of classes in the dataset.
$l_t[p][c]$ (short for $l_t[p_y][p_x][c]$) indicates the model's confidence that $p$ in tile $t$
belongs to class $c \in C$.

Then the class $c_p$ of $p$ in the merged segmentation map can be given by:
\begin{equation}
    c_p = \arg\max_{c \in C} \left( \max_{t \in T} l_t[p][c] \right)\label{eq:prediction-merge}.
\end{equation}

In other words, $c_p$ maximizes the maximum logit over all tiles overlapping at $p$.

\begin{landscape}
    \thispagestyle{empty}
    \begin{minipage}[H][\textheight][s]{\linewidth}
    \subsection{Merging predictions by cropping tiles}\label{subsec:appendix:crop-merging}
    An alternative approach that is faster and yields higher quality results is to crop tiles by half
    the overlap factor (stride) along each side of the tile.
    Intuitively, the model has the most information about objects in the center of a tile.
    The further towards the border an object lies, the more likely it is to be cut off and not
    be identified accurately.\\ [1.5ex]
    Programmatically, the cropping approach is faster because it can utilize vectorized array slicing operations
    to copy large chunks of each tile into the final segmentation map, whereas the previous approach
    performs per-pixel operations in plain Python code, making it orders of magnitude slower.
    Moreover, model confidence as expressed through logits can be noisy or erronous, and may therefore
    not be the best merging criteria, particularly when the model is under-trained.
    Fig.~\ref{fig:prediction-merge-methods} illustrates these differences with an example.
    \begin{figure}[H]
        \centering
        \begin{subfigure}[b]{0.49\linewidth}
            \includegraphics[width=\linewidth]{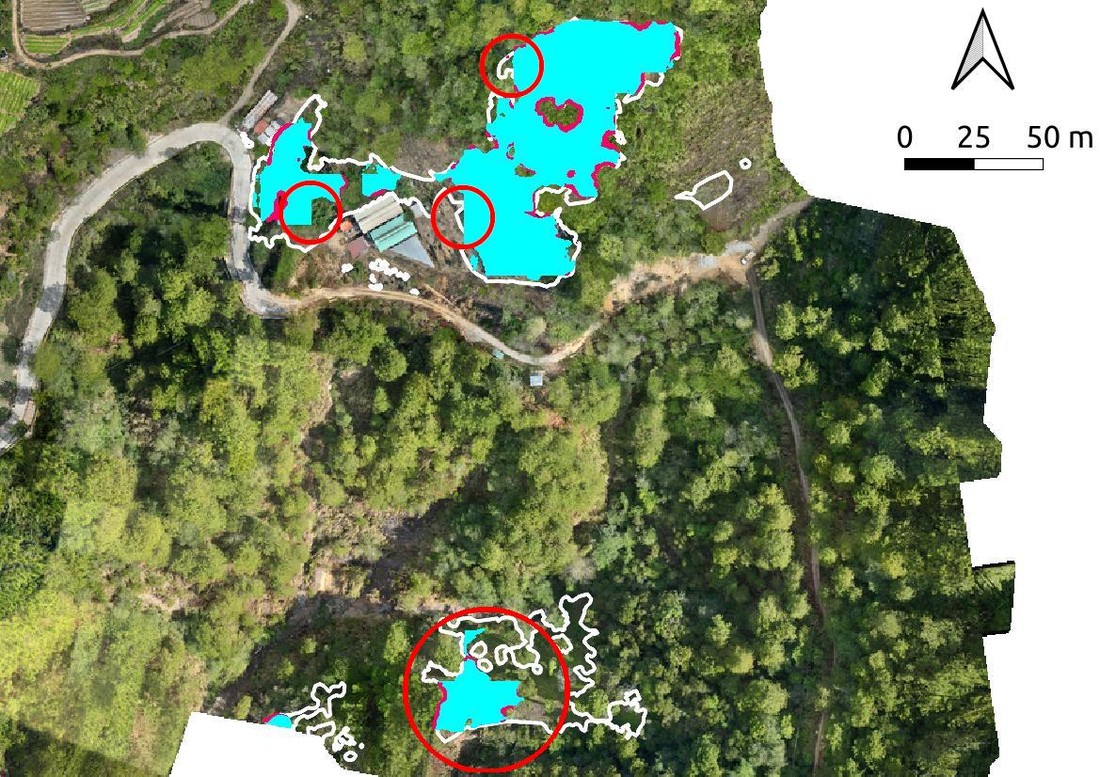}
            \caption{}
        \end{subfigure}
        \hfill
        \begin{subfigure}[b]{0.49\linewidth}
            \includegraphics[width=\linewidth]{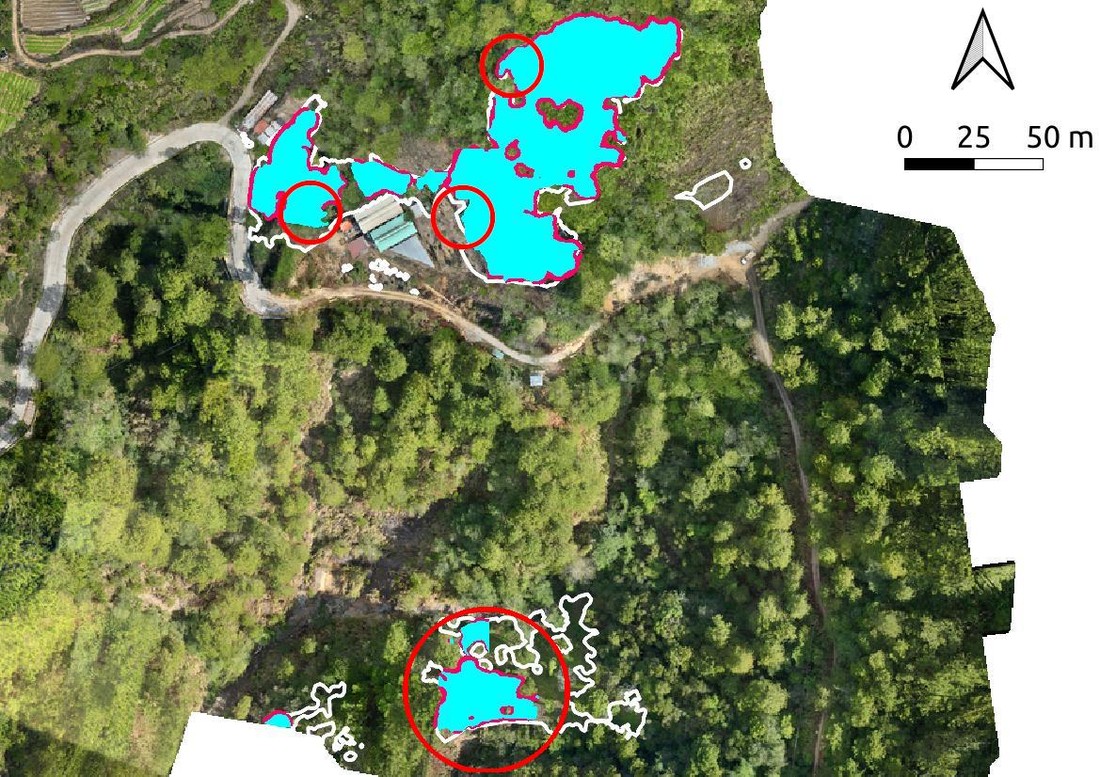}
            \caption{}
        \end{subfigure}

        \caption{Comparison of max-logit merging (a) vs tile-crop merging (b) for Chayote
        predictions at 0.3 m/px. Merging by cropping reduces artifacts and substantially increases
        the border size. Red circles indicate locations where the improvement is particularly
        visible.}
        \label{fig:prediction-merge-methods}
    \end{figure}
    \fillandplacepagenumber
    \end{minipage}
\end{landscape}

\newpage

\section{Overview of resolutions}\label{sec:appendix:resolutions}

\begin{table}[ht]
    \caption{Sto.\ Niño dataset resolutions and matching platforms}
    \centering
    \begin{tabular}{cccc}
        \hline\hline
        \rule{0pt}{3ex}
        GSD   & Resolution (W) & Resolution (H) & Platform \\ [1ex]
        \hline
        \rule{0pt}{3ex}
        0.08  & 23662          & 25228          & UAV                       \\
        0.10  & 18930          & 20182          & UAV                       \\
        0.12  & 15775          & 16819          & UAV                       \\
        0.15  & 12620          & 13455          & Maxar (Upscaled)          \\
        0.30  & 6310           & 6727           & Maxar, Digital Globe      \\
        0.50  & 3786           & 4036           & Maxar, Airbus             \\
        0.70  & 2704           & 2883           & Maxar, Planet, CNES, KARI \\
        1     & 1893           & 2018           & Lockheed Martin Space     \\
        3     & 631            & 673            & Planet                    \\
        5     & 379            & 404            & RapidEye Blackbridge      \\
        10    & 189            & 202            & Sentinel-2                \\
        15    & 126            & 135            & Landsat               \\ [1ex]
        \hline
    \end{tabular}
    \label{tab:resolutions-overview}
\end{table}

\section{Detailed discussion of degradation methods}\label{sec:appendix:degradation-methods}
The degradation methods of this study were illustrated in Fig.~\ref{fig:resolution-degradation}.
Method A downscales images and labels during training before they are aggregated
into a mini-batch.
This reduces the amount of information per input to the model, while maintaining
pixel density, thus each pixel represents a larger physical distance.
It also requires a different model architecture to accommodate for the new input size, which
makes comparisons to the baseline more difficult.
Downscaling of labels reduces the precision of gradient computations when updating
model weights, which impedes training progress at low resolutions.
\verb|INTER_AREA| interpolation of the OpenCV library~\cite{Bradski2000} was used for
downscaling.

Pixelation (method B) does not affect input size, but also reduces the amount of conveyed
information, while lowering pixel density.
Unlike method A, labels are not degraded, which gives a better estimate of how image quality alone
affects predictions.
Method B represents a synthetic study of resolution degradation, as multiple pixels in the
final tile represent a single pixel in the downsized tile.
\verb|INTER_AREA| interpolation was again used for downscaling,
followed by bicubic upscaling.

Method C creates tiles which are ``zoomed out'' compared to the original input~\cite{Barbedo2019}.
Detailed information at the pixel level is lost, but the model also receives a larger view of the
surrounding area.
The pixel density in this approach is unaffected, and each pixel covers a larger spatial distance.
Most importantly, method C accurately simulates the change per tile when the
satellite or drone capturing the imagery has higher altitude or is equipped with a sensor
registering fewer samples.
However, as the resolution of the orthomosaic is lowered, fewer tiles can be produced.
This reduction in training data can affect model performance and lead to overfitting, and is
investigated in detail in section~\ref{sec:evaluation}.
Each degraded dataset was generated by downscaling the original dataset using lanczos interpolation
in QGIS~\cite{qgis2023}.

\section{Mask2Former}\label{sec:appendix:mask2former}
Several prerequisites are needed to understand transformer networks, as they build on years of
research into natural language processing (NLP) and input encoding.
For brevity, only a brief overview is provided here, but readers are encouraged to refer to both
the original paper on transformers, ``Attention is all you need''~\cite{Vaswani2023}, as well the
excellent video series ``StatQuest'' by Josh
Starmer\footnote{\url{https://www.youtube.com/watch?v=zxQyTK8quyY}}.

\subsection{Transformer neural networks and the self attention mechanism}
The transformer architecture is proving to outperform all previous approaches in NLP, computer
vision (CV), data generation, and more.
Its strong performance is not attributed to smart design choices in model architecture, though.
In fact, the transformer is a far more general model than CNNs, LSTMs, and even MLPs.

Prior model architectures across domains heavily relied on inductive biases.
Such biases tell the model \textit{how} to tackle a task, and through training the model becomes
better at applying the prescribed technique to the data, be it convolutions for image analysis,
long short-term memory for text prediction, etc.
Crucially, because the methodology is already defined, far less training data is needed for the
model to converge to a trained state.

In contrast, the transformer architecture does not specify how a task should be solved.
Inputs to the model are encoded to register their semantic meaning and position in the sequence,
and using a mechanism called self-attention, the similarity between a sequence token to
itself and all other tokens is computed.
This self attention computation is quadratic in runtime, but can be computed in parallel for all
tokens, which makes transformers far more computationally efficient than previous architectures.

However, the strong reduction in inductive biases require an abundance of training data for
the model to converge.
Not only must relationships in the data be identified by the model, but analysis techniques must
first be learned.
By allowing the model to develop its own methodology, with enough training data and time it can
derive approaches that outperform ``fixed'' architectures like CNNs.
Interestingly, it was shown that transformers could independently develop techniques akin to CNN
convolution for CV tasks, and hierarchical analysis of grammar, context, and meaning in NLP tasks.

Conventional model architectures will still retain relevance whenever computational requirements
or data availability are constrained.
For the same reason, many popular transformer implementations are first pretrained on large generic
datasets, such as ImageNet, to allow transfer training of models to specific tasks with fewer
computational resources.

\subsection{Vision transformers}
The vision transformer (ViT) was first proposed in 2020~\cite{Dosovitskiy2020}.
At the time, transformers in computer vision (CV) were employed by combining parts of their model architecture with CNNs.
This is because self attention is prohibitively expensive to compute already for long text
sequences, but images typically contain orders of magnitude more data points than text.

\noindent The authors in~\cite{Dosovitskiy2020} proposed a ``pure'' transformer architecture for CV\@.
They solved the computational issue of self attention on images by first dividing the input image
into so-called patches of 16x16.
Using a linear transformation, each patch (unrolled to a 1-d vector) is projected into a lower
dimensional space, and the resulting vector is positionally embedded, producing a patch embedding.
Patch embeddings are then fed to a regular transformer.
Learnable \verb|[class]| embeddings are fed into the transformer encoder as the final inputs of the
sequence, to produce an output that is fed to an MLP head for image classification.

The ViT architecture therefore solves the computational overhead of self attending to images, while
still benefiting from the parallelization capabilities of transformers.
This enabled the authors to train their ViT on orders of magnitude more data, and outperform all
prior CNN-based approaches.

\subsection{MaskFormer and Mask2Former}
Semantic segmentation tasks, regardless of model architecture, were largely tackled as per-pixel
classification problems.
With the introduction of ViTs, works such as~\cite{strudel2021segmenter} and~\cite{zheng2021rethinking}
demonstrated pixel-level semantic segmentation using transformer encoders and custom
decoders that upsample the encoded input and transform it into the segmentation result.

However, pixel-level segmentation is not the only approach to semantic segmentation.
Mask classification is an alternative approach where the model produces a number of segmentation
masks for the same image, with each mask having a single class ID.
While this enables semantic segmentation, it also allows for instance segmentation, where objects
within the same class are distinguished.

MaskFormer~\cite{cheng2021perpixel} was introduced as the first ViT model to use mask classification
for image segmentation.
Its architecture naturally allows for semantic and instance segmentation.
Moreover, models performing per-pixel semantic segmentation can be converted to the MaskFormer
architecture, because MaskFormer uses a pixel-level module to generate feature maps from input
images.

Mask2Former~\cite{cheng2021mask2former} later followed to improve upon the MaskFormer architecture.
Its main difference is the use of masked attention, where unlike a regular transformer decoder
used in MaskFormer, the decoder is restricted to focus on areas near predicted segments.
Further, memory usage was optimized by computing loss over a set of randomly selected points, rather
than entire predictions.

\newpage

\section{Numeric data on per-feature model performance relative to survey characteristics}\label{sec:appendix:degradation-scores}
\begin{table}[ht]
    \caption{Quantative evaluation of Chayote segmentation performance as GSD is increased}
    \label{tab:chayote-degradation-scores}
    \resizebox{\textwidth}{!}{%
        \begin{tabular}{cccccccc}
            \hline\hline
            \rule{0pt}{3ex}
            Method &
            \multicolumn{1}{l}{GSD (m/px)} &
            \multicolumn{1}{l}{mIoU} &
            \multicolumn{1}{l}{mDice} &
            \multicolumn{1}{l}{IoU Background} &
            \multicolumn{1}{l}{IoU Chayote} &
            \multicolumn{1}{l}{IoU Border} &
            \multicolumn{1}{l}{Best val. iteration} \\ [1ex]
            \hline
            \rule{0pt}{3ex}
            n/a                 & 0.08 & 66.79 & 74.83 & 98.64 & 79.2  & 22.53 & 14500 \\ [1ex]
            \hline
            \rule{0pt}{3ex}
            \multirow{11}{*}{A} & 0.10 & 67.13 & 75.34 & 98.64 & 78.94 & 23.81 & 6500  \\
            & 0.12 & 64.85 & 73.84 & 98.24 & 72.67 & 23.65 & 2500  \\
            & 0.15 &  58.89 & 67.56 & 97.73 & 64.24 & 14.68 & 18500 \\
            & 0.30 & 55.94 & 64.22 & 97.39 & 59.98 & 10.53 & 11500 \\
            & 0.50 & 46.80 & 54.80 & 96.23 & 38.47 & 5.69  & 16500  \\
            & 0.70 & 42.07 & 49.53 & 93.96 & 28.28 & 3.96  & 7500 \\
            & 1    & 41.79 & 49.36 & 94.23 & 26.02 & 5.13  & 11500 \\
            & 3    & 34.43 & 38.39 & 92.86 & 10.43 & 0     & 6500 \\
            & 5    & 32.37 & 34.25 & 94.12 & 2.98  & 0     & 19000 \\
            & 10   & 31.47 & 32.37 & 94.4  & 0     & 0     & 500   \\
            & 15   & 31.5  & 32.39 & 94.5  & 0     & 0     & 500   \\ [1ex]
            \hline
            \rule{0pt}{3ex}
            \multirow{11}{*}{B} & 0.10 & 64.68 & 72.7  & 98.36 & 76.42 & 19.26 & 13500 \\
            & 0.12 & 63.80 & 72.09 & 98.34 & 73.95 & 19.11 & 18000  \\
            & 0.15 & 60.84 & 69.72 & 97.84 & 67.10 & 17.60 & 21500 \\
            & 0.30 & 61.31 & 70.11 & 97.97 & 68.04 & 17.92 & 8500  \\
            & 0.50 & 54.75 & 64.18 & 97.11 & 53.07 & 14.07 & 13000 \\
            & 0.70 & 50.59 & 59.76 & 96.52 & 44.42 & 10.82 & 11500  \\
            & 1    & 48.84 & 57.85 & 96.15 & 40.65 & 9.72  & 7500  \\
            & 3    & 42.17 & 49.53 & 94.59 & 27.86 & 4.05  & 11500 \\
            & 5    & 35.77 & 40.71 & 92.34 & 14.78 & 0.19  & 11500  \\
            & 10   & 33.16 & 38.05 & 88.19 & 10.85 & 0.43  & 15500 \\
            & 15   & 32.66 & 9.16  & 88.83 & 9.16  & 0     & 13500 \\ [1ex]
            \hline
            \rule{0pt}{3ex}
            \multirow{11}{*}{C} & 0.10 & 65.78 & 73.97  & 98.46 & 77.33 & 212.56 & 8500 \\
            & 0.12 & 65.62 & 73.48 & 98.60 & 78.24 & 20.02 & 6500 \\
            & 0.15 & 64.11 & 72.81 & 98.34 & 72.79 & 21.21 & 2500  \\
            & 0.30 & 56.73 & 66.34 & 97.6  & 56.04 & 16.54 & 1528  \\
            & 0.50 & 50.66 & 60.12 & 96.96 & 42.49 & 12.53 & 630   \\
            & 0.70 & 42.56 & 49.26 & 95.76 & 29.99 & 1.94  & 342   \\
            & 1    & 42.44 & 48.88 & 96.74 & 28.73 & 1.86  & 90    \\
            & 3    & 32.99 & 33.16 & 98.97 & 0     & 0     & 10    \\
            & 5    & 33.21 & 33.27 & 99.64 & 0     & 0     & 6     \\
            & 10   & 33.3  & 33.32 & 99.91 & 0     & 0     & 5     \\
            & 15   & 33.32 & 33.33 & 99.96 & 0     & 0     & 3        \\ [1ex]
            \hline
        \end{tabular}%
    }
\end{table}
\begin{table}[H]
    \caption{Quantative evaluation of road segmentation performance as GSD is increased}
    \label{tab:road-degradation-scores}
    \resizebox{\textwidth}{!}{%
        \begin{tabular}{cccccccc}
            \hline\hline
            \rule{0pt}{3ex}
            Method &
            GSD (m/px) &
            \multicolumn{1}{l}{mIoU} &
            \multicolumn{1}{l}{mDice} &
            \multicolumn{1}{l}{IoU Background} &
            \multicolumn{1}{l}{IoU Roads} &
            \multicolumn{1}{l}{Best val. iteration} \\ [1ex]
            \hline
            \rule{0pt}{3ex}
            n/a                 & 0.08 & 79.84 & 87.49 & 99.16 & 60.52 & 5000 \\ [1ex]
            \hline
            \rule{0pt}{3ex}
            \multirow{11}{*}{A} & 0.10 & 77.85 & 85.92 & 99.08 & 56.62 & 3000  \\
            & 0.12 & 77.39 & 85.54 & 99.08 & 55.71 & 9000  \\
            & 0.15 & 75.92 & 84.32 & 98.99 & 52.84 & 14000 \\
            & 0.3  & 72.47 & 81.25 & 98.87 & 46.07 & 2000  \\
            & 0.5  & 66.21 & 74.93 & 98.57 & 33.85 & 9000  \\
            & 0.7  & 65.59 & 74.27 & 98.44 & 32.73 & 9000  \\
            & 1    & 67.36 & 76.21 & 98.51 & 36.21 & 4000  \\
            & 3    & 51.01 & 53.34 & 98.03 & 3.99  & 15000 \\
            & 5    & 49.05 & 49.52 & 98.09 & 0.00  & 500   \\
            & 10   & 48.95 & 49.47 & 97.90 & 0.00  & 500   \\
            & 15   & 48.96 & 49.48 & 97.92 & 0.00  & 500   \\ [1ex]
            \hline
            \rule{0pt}{3ex}
            \multirow{11}{*}{B} & 0.10 & 78.35 & 86.32 & 99.09 & 57.61 & 3000  \\
            & 0.12 & 78.18 & 86.18 & 99.11 & 57.25 & 3000  \\
            & 0.15 & 76.24 & 84.59 & 99.02 & 53.46 & 3000  \\
            & 0.3  & 78.10 & 86.12 & 99.09 & 57.12 & 3500  \\
            & 0.5  & 73.64 & 82.32 & 98.95 & 48.34 & 16500 \\
            & 0.7  & 73.40 & 82.10 & 98.92 & 47.84 & 4000  \\
            & 1    & 73.10 & 81.86 & 98.82 & 47.39 & 4000  \\
            & 3    & 63.75 & 72.12 & 98.45 & 29.05 & 4000  \\
            & 5    & 59.69 & 66.98 & 98.31 & 21.08 & 22500 \\
            & 10   & 57.19 & 63.67 & 97.80 & 16.58 & 15500 \\
            & 15   & 55.30 & 60.71 & 97.95 & 12.64 & 15500 \\ [1ex]
            \hline
            \rule{0pt}{3ex}
            \multirow{11}{*}{C} & 0.10 & 79.86 & 87.50 & 99.20 & 60.52 & 6500  \\
            & 0.12 & 79.77 & 87.44 & 99.15 & 60.38 & 2500  \\
            & 0.15 & 79.50 & 87.22 & 99.21 & 59.79 & 2500  \\
            & 0.3  & 74.57 & 83.13 & 99.12 & 50.03 & 1146  \\
            & 0.5  & 72.21 & 80.98 & 99.04 & 45.38 & 140   \\
            & 0.7  & 72.99 & 81.72 & 99.00 & 46.99 & 152   \\
            & 1    & 68.45 & 77.24 & 98.97 & 37.93 & 54    \\
            & 3    & 49.83 & 49.92 & 99.66 & 0.00  & 2     \\
            & 5    & 49.79 & 52.13 & 96.45 & 3.13  & 7     \\
            & 10   & 50.64 & 51.38 & 99.84 & 1.44  & 8     \\
            & 15   & 50.00 & 50.00 & 99.99 & 0.00  & 1     \\ [1ex]
            \hline
        \end{tabular}
    }
\end{table}

\begin{table}[H]
    \caption{Quantative evaluation of feature segmentation performance as train dataset size is decreased}
    \label{tab:area-reduction-scores}
    \resizebox{\textwidth}{!}{%
        \begin{tabular}{ccccccc}
            \hline\hline
            \rule{0pt}{3ex}
            Feature                   & Tiles in Train Set & mIoU  & mDice & IoU Background & IoU Feature & IoU Border \\ [1ex]
            \hline
            \rule{0pt}{3ex}
            \multirow{10}{*}{Chayote} & 3001               & 66.79 & 74.83 & 98.64          & 79.2        & 22.53      \\
            & 1963               & 67.52 & 75.81 & 98.65          & 79.1        & 24.82      \\
            & 1388               & 68.34 & 76.37 & 98.71          & 81.18       & 25.11      \\
            & 910                & 65.55 & 73.81 & 98.36          & 76.81       & 21.49      \\
            & 254                & 56.68 & 67.02 & 96.93          & 53.25       & 19.85      \\
            & 93                 & 55.31 & 66.03 & 96.86          & 47.99       & 21.09      \\
            & 50                 & 41.69 & 48.94 & 95.31          & 24.79       & 4.97       \\
            & 23                 & 32.37 & 34.93 & 92.61          & 4.28        & 0.21       \\
            & 2                  & 28.67 & 31.66 & 84.21          & 1.37        & 0.43       \\
            & 1                  & 31.03 & 32.39 & 92.55          & 0.51        & 0.01       \\ [1ex]
            \hline
            \rule{0pt}{3ex}
            \multirow{10}{*}{Roads}   & 2781               & 79.84 & 87.49 & 99.16          & 60.52       & n/a        \\
            & 1836               & 77.26 & 85.43 & 99.12          & 55.39       & n/a        \\
            & 1304               & 74.17 & 82.78 & 99.01          & 49.32       & n/a        \\
            & 862                & 77.2  & 85.38 & 99.12          & 55.28       & n/a        \\
            & 246                & 68.57 & 77.46 & 98.65          & 38.5        & n/a        \\
            & 101                & 63.58 & 71.88 & 98.56          & 28.6        & n/a        \\
            & 56                 & 58.34 & 59.85 & 98.37          & 18.32       & n/a        \\
            & 31                 & 55.8  & 61.37 & 98.19          & 13.41       & n/a        \\
            & 4                  & 49.08 & 49.54 & 98.17          & 0           & n/a        \\
            & 1                  & 49.08 & 49.54 & 98.17          & 0           & n/a        \\ [1ex]
            \hline
        \end{tabular}
    }
\end{table}

\begin{landscape}
    \section{Degradation visualization}\label{sec:appendix:degradation-visualization}
    \begin{figure}[ht]
        \centering
        \begin{subfigure}[b]{0.49\linewidth}
            \includegraphics[width=\linewidth]{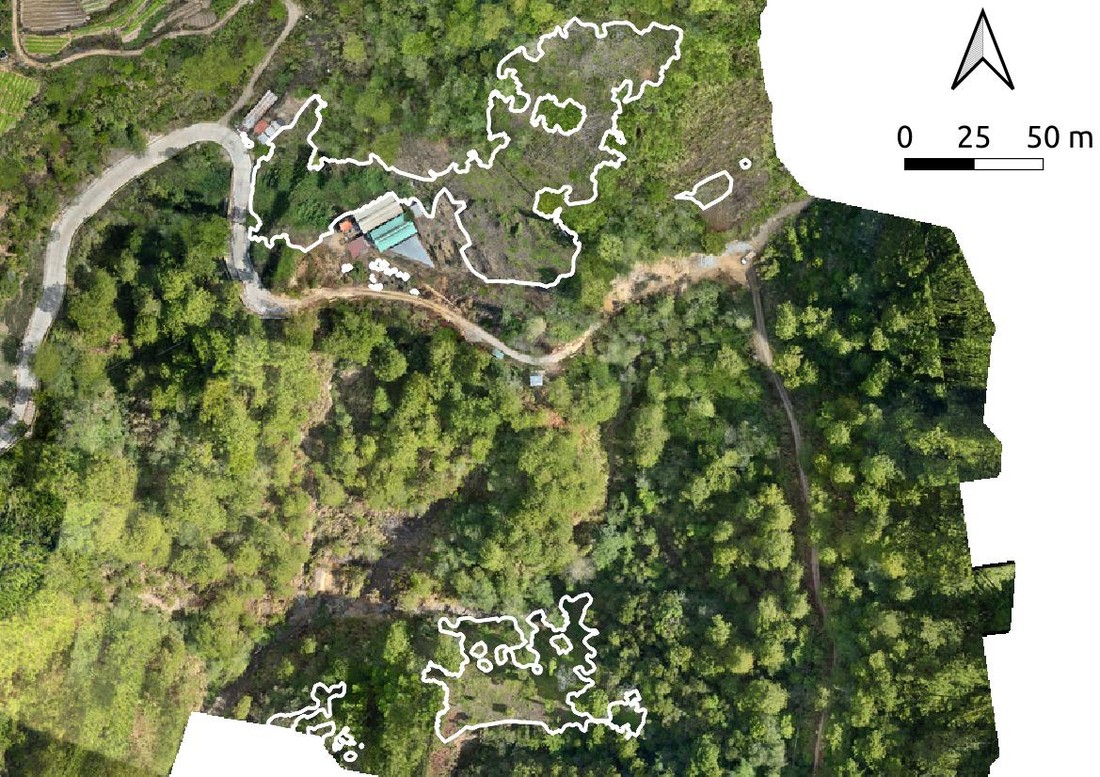}
            \caption{}
        \end{subfigure}
        \begin{subfigure}[b]{0.49\linewidth}
            \includegraphics[width=\linewidth]{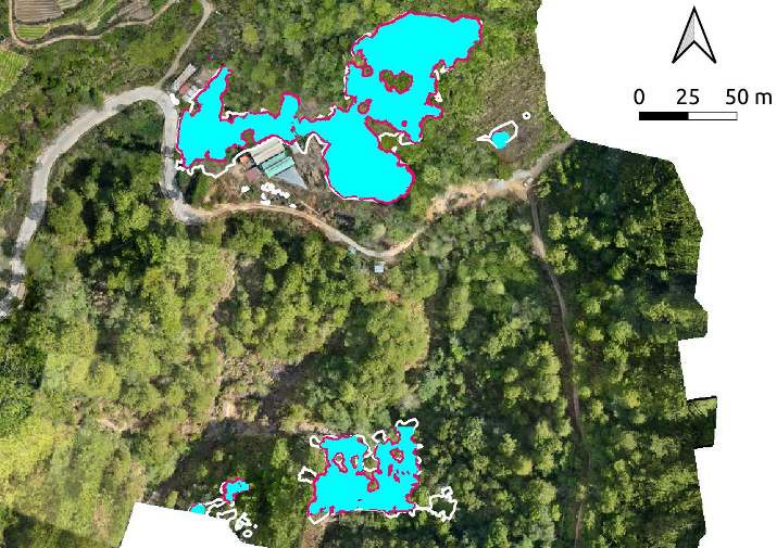}
            \caption{}
        \end{subfigure}
        \caption{Setup for visualization of performance degradation for Chayote w.r.t. GSD.
            (a) Ground truth Chayote labels (border merged); and (b) model prediction at
            8cm/px. Cyan and red indicate predicted Chayote and border, respectively.
            See Fig.~\ref{fig:degradation-visually} for degradation over each method.}
        \label{fig:degradation-info}
    \end{figure}
    \fillandplacepagenumber
\end{landscape}
\newpage
\begin{landscape}
    \thispagestyle{empty}
    \begin{figure}[ht]
        \centering
        \includegraphics[width=\linewidth]{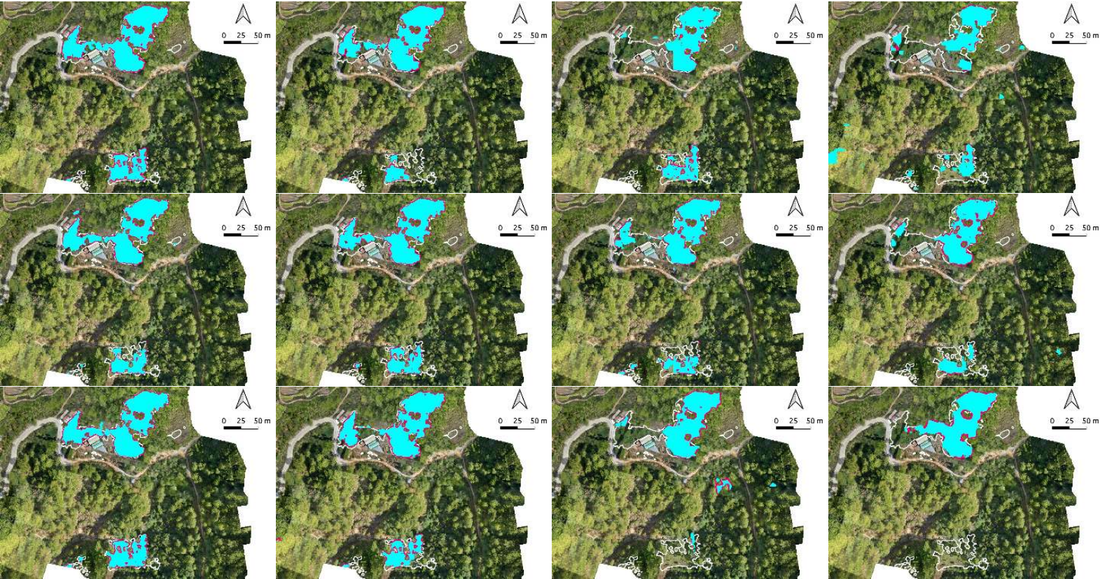}
        \caption{Visualization of performance degradation on Chayote with decreasing GSD.
        Degradation methods: A, top row; B, center row; C, bottom row.
        GSDs per row in m/px, from left to right: 0.10, 0.15, 0.30, 0.50.
        Starting from 0.3 m/px, performance starts to degrade across all methods, whereas
        segmentation peformance going from 0.10 m/px to 0.15 m/px is only affected less.}
        \label{fig:degradation-visually}
        \fillandplacepagenumber
    \end{figure}
\end{landscape}

\section{Exemplary applications of the Cording Index}\label{sec:appendix:ca-examples}

\subsection{GSD prediction for vineyard leave density estimation}
\cite{Ilniyaz2022} studied the ability of multiple machine learning models such as support vector
regression and k-nearest neighbor to estimate the leaf area index (LAI), which measures the total
leaf area for a given ground area, indicating the density of leaves.
The study was conducted over a small vineyard in Turpan city, Xinjang, China, where (according
to~\cite{Ilniyaz2022}) the most widespread grape vine variety is Vitis vinifera.

The authors found that for two GSDs, 0.007 and 0.045 m/px, the $R^2$ values of their best
model were 0.825 and 0.637, respectively, when using RGB imagery to estimate LAI.

The leaves of the Vitis vinifera variety are heart-shaped, and approximately 5--15 cm
wide~\cite{NPTVitis, NCEGPTVitis}.
\metric then produces the interval $(\frac{0.05}{3}\approx0.016,\frac{0.15}{3}=0.05)$.
Based on the two provided $R^2$ values, this is a sensible range for where performance
deteriorates.

\subsection{GSD prediction for cows and sheep}
In~\cite{Brown2022} it was found that mean average precision (mAP) for detecting cows dropped
sharply starting around 0.15--0.20 m/px.
Although the species of cow was not specified, average body width of common cow breeds lies
between 51.7--69 cm~\cite{Cerqueira2013, Santana_2022a, Santana_2022b, ontario}.
Many cows in the study had black fur~\cite{Brown2022}, so their bodies (rather than fur texture)
were chosen as SVA\@.
The SVA was measured as body width because it is the shortest axis for cows in overhead imagery.

\metric then gives a GSD interval of $(\frac{0.517}{3}\approx0.172,\frac{0.69}{3}=0.23)$,
which matches the observed start of performance degradation well.
Additionally, GSDs beyond 0.23 resulted in mAPs of 0.6 or lower, from the initial score of about
0.95, so the metric also provides a sensible cutoff point.

The mAP for sheep in~\cite{Brown2022} was initially more stable for three of the four employed
degradation methods, remaining close to 1, decreasing at a GSD around 0.2 m/px and dropping sharply
at 0.3 m/px.
For the circular degradation method with input size 1280x1280 (``1280 Circular''), performance
decreased slightly at a GSD of 0.1 m/px, and dropped sharply at 0.2 m/px.

Width of domestic sheep is not well documented, but~\cite{Omondi_2022} reports a body width of
44--66 cm.
The interval derived via \metric is $(\frac{0.44}{3}\approx0.147,\frac{0.66}{3}=0.22)$.
This estimate suits the observed degradation trends well.
In particular, the lower bound lies before significant decreases
across all four degradation methods, and the upper bound is tight enough to provide a suitable
cutoff also for the 1280 Circular degradation method.

\section{Effect of target pixel density on survey characteristics}\label{sec:appendix:pixel-density-area}
\begin{figure}[H]
    \centering
    \begin{subfigure}[b]{0.45\textwidth}
        \includegraphics[width=\textwidth]{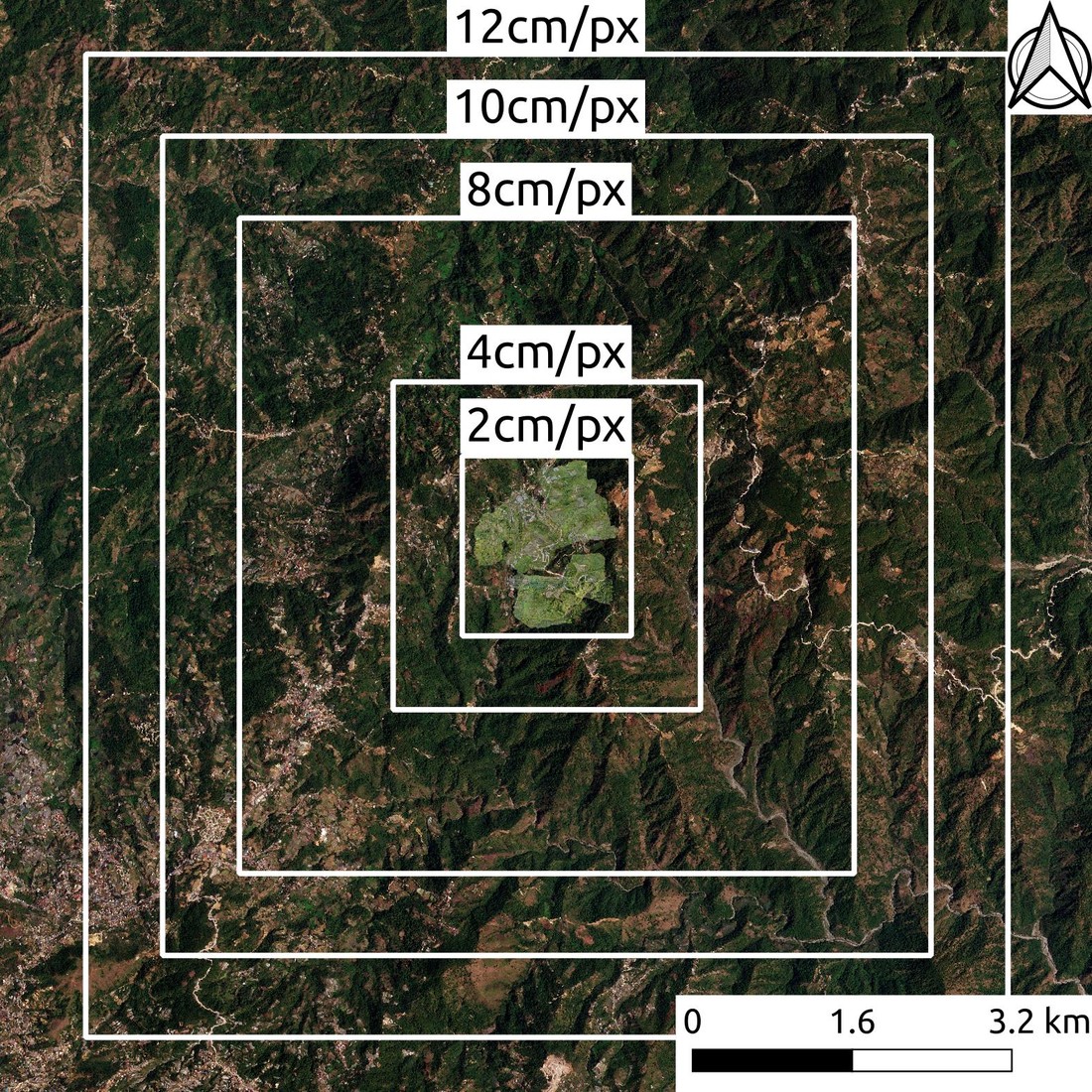}
        \caption{}
        \label{fig:resolution_area_1}
    \end{subfigure}
    \begin{subfigure}[b]{0.45\textwidth}
        \includegraphics[width=\textwidth]{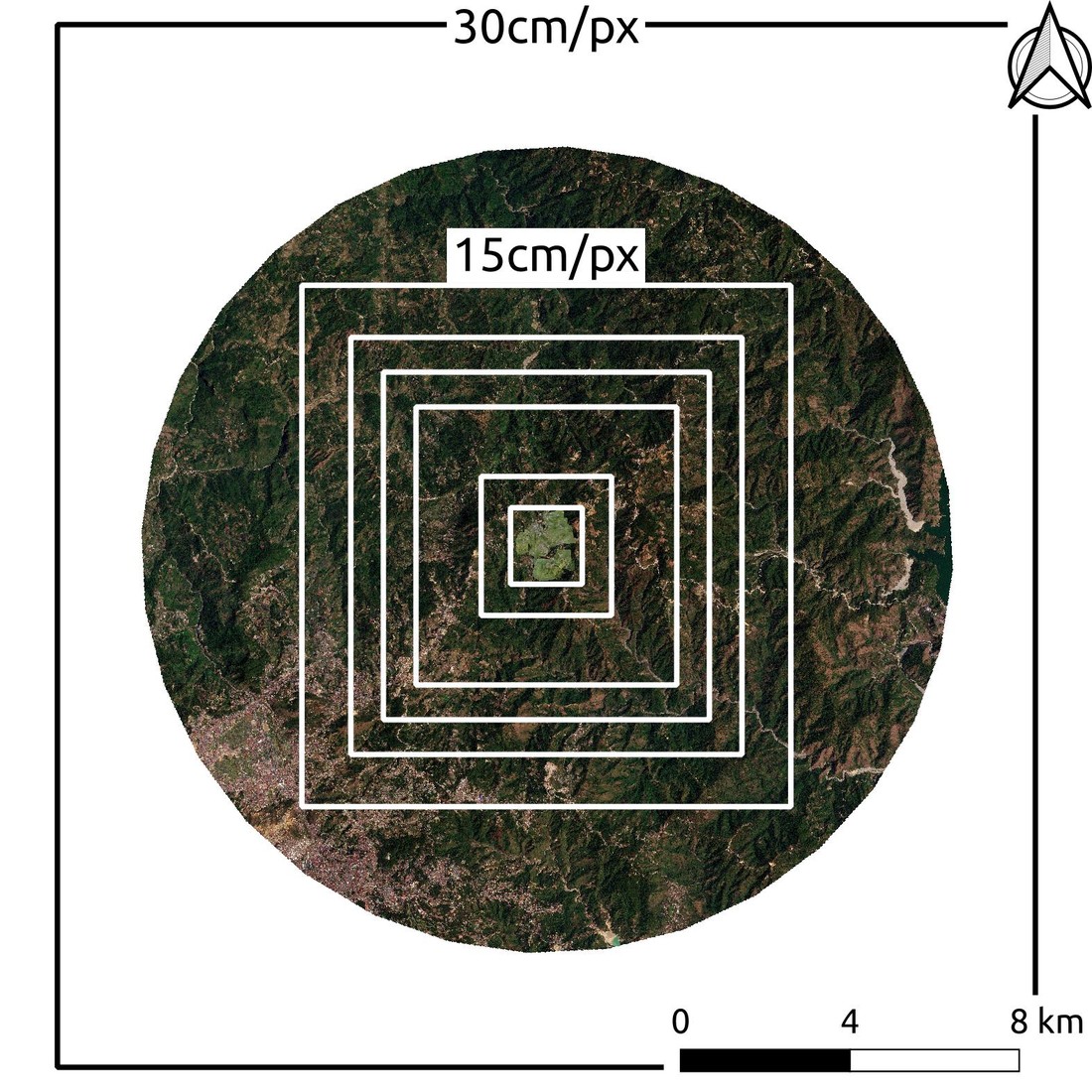}
        \caption{}
        \label{fig:resolution_area_2}
    \end{subfigure}
    \caption{
        Change in geographic extent relative to spatial resolution when fixing image dimension (in
        pixels) to native dataset resolution at 0.022 m/px. (a) Sto.\ Niño dataset overlaid
        onto Planet satellite imagery, with bounding boxes for 5 VHR GSDs; (b)
        additional extents for 0.15 and 0.3 m/px GSDs. The satellite basemap covers
        an area of approximately 286.27 km$^2$. Extents for remaining resolutions above
        0.30m/px fall outside the figure.
    }
    \label{fig:resolution-area}
\end{figure}
\begin{table}[H]
    \centering
    \caption{
        Estimated flight statistics for a DJI Mavic 3M drone imaging each region shown in
        Fig.~\ref{fig:resolution-area}, such that the number of pixels in each orthomosaic is the
        same as for the native dataset with a GSD of 0.022 m/px.
        Statistics for GSDs of 3, 5, 10, and 15 m/px not shown as the maximum flight altitude
        of 6km would be exceeded.
        The maximum legal flight height in most countries is 120 metres, thus a lower quality sensor
        would need to be fitted on the drone, rather than increasing altitude as shown here.
        Calculations are based on the specifications provided by DJI.
    }
    \label{tab:dji-stats}
    \resizebox{\textwidth}{!}{%
        \begin{tabular}{ccccc}
            \hline\hline
            \rule{0pt}{3ex}
            GSD (m/px) & Altitude (m) & Area to image (km$^2$) & \begin{tabular}[c]{@{}c@{}}8h workdays to complete imaging\\ (optimistic estimates)\end{tabular} \\ [0.5ex]
            \hline
            \rule{0pt}{3ex}
            0.022               & 75.08                 & 3.06                            & 0.30                                     \\
            0.04                & 150.16                & 10.12                           & 1.34                                     \\
            0.08                & 300.32                & 40.48                           & 2.22                                     \\
            0.10                & 375.40                & 63.26                           & 4.44                                     \\
            0.12                & 450.48                & 91.09                           & 5.55                                     \\
            0.15                & 563.098               & 142.32                          & 6.66                                     \\
            0.30                & 1126.20               & 569.30                          & 8.33                                     \\
            0.50                & 1876.99               & 1581.38                         & 16.66                                    \\
            0.70                & 2627.79               & 3099.50                         & 27.76                                    \\
            1.00                & 3753.99               & 6325.51                         & 38.87                                    \\ [1ex]
            \hline
        \end{tabular}
    }
\end{table}

\newpage
\section{Example of \product applicability}\label{sec:appendix:example-of-product-applicability}
\begin{figure}[ht]
    \includegraphics[width=\textwidth]{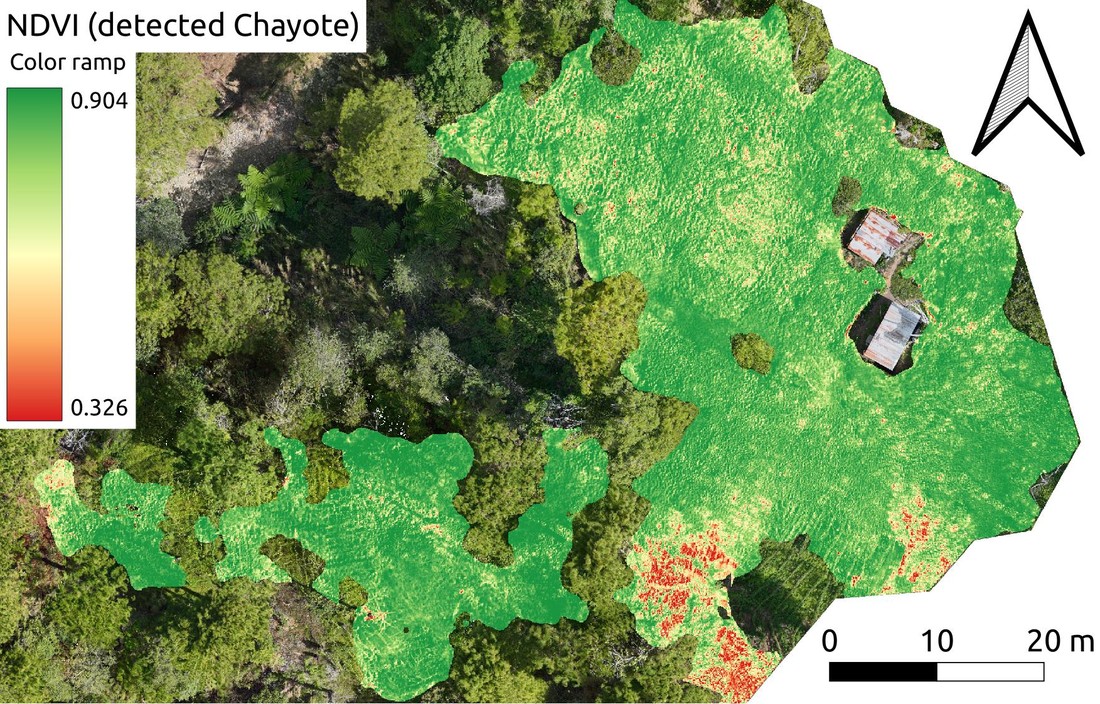}
    \caption{Normalized difference vegetation index (NDVI) for a Chayote plantation in the
    test set, using the model's predictions as mask.
    The NDVI indicates plant health and lies between -1 and 1. By masking the NDVI map,
        it shows the health of Chayote plants relative to each other, which helps identify which
        areas need irrigation or treatment.}
    \label{fig:ndvi}
\end{figure}

\end{document}